\def\year{2022}\relax
\documentclass[letterpaper]{article} 
\usepackage{aaai22}  
\usepackage{times}  
\usepackage{helvet}  
\usepackage{courier}  
\usepackage[hyphens]{url}  
\usepackage{graphicx} 
\usepackage{natbib}  
\usepackage{caption} 
\DeclareCaptionStyle{ruled}{labelfont=normalfont,labelsep=colon,strut=off} 
\usepackage[utf8]{inputenc}
\usepackage{hyperref}       
\usepackage{booktabs}       
\usepackage{amsfonts}       
\usepackage{nicefrac}       
\usepackage{microtype}      
\usepackage{xcolor}         
\usepackage{microtype}
\usepackage{graphicx}
\usepackage{subfigure}
\usepackage{booktabs} 

\usepackage{amsmath,amssymb,amsthm,color,graphicx,mathrsfs,booktabs,algorithm,algorithmic,bm,comment,bbm,wrapfig}
\newtheorem{theorem}{Theorem}
\newtheorem{proposition}{Proposition}

\newtheorem{corollary}{Corollary}
\newtheorem{lemma}{Lemma}

\DeclareMathOperator*{\argmin}{arg\,min}

\allowdisplaybreaks

\title{Hypergraph Modeling via Spectral Embedding Connection:\\
Hypergraph Cut, Weighted Kernel $k$-means, and Heat Kernel}
\author {
    Shota Saito
}
\affiliations{
    Department of Computer Science,
    University College London\\
    ssaito@cs.ucl.ac.uk
}
\begin{document}
\maketitle
\begin{abstract}
We propose a theoretical framework of multi-way similarity to model real-valued data into hypergraphs for clustering via spectral embedding.
For graph cut based spectral clustering, it is common to model real-valued data into graph by modeling pairwise similarities using kernel function.
This is because the kernel function has a theoretical connection to the graph cut.
For problems where using multi-way similarities are more suitable than pairwise ones, it is natural to model as a hypergraph, which is generalization of a graph.
However, although the hypergraph cut is well-studied, there is not yet established a hypergraph cut based framework to model multi-way similarity.
In this paper, we formulate multi-way similarities by exploiting the theoretical foundation of kernel function.
We show a theoretical connection between our formulation and hypergraph cut in two ways, generalizing both weighted kernel $k$-means and the heat kernel, by which we justify our formulation.
We also provide a fast algorithm for spectral clustering.
Our algorithm empirically shows better performance than existing graph and other heuristic modeling methods.
\end{abstract}
\section{Introduction}
Graphs are widely used data representations for data that have pairwise relationships.
One of the main aims for graph machine learning is clustering vertices, and the graph cut based spectral clustering is a popular method~\cite{normcut,tuto}.
For clustering purposes, spectral clustering is also useful for real-valued data.
We model real-valued data as a graphs by forming a vertex from each data point and an edge from pairwise similarity of each pair of data points~\cite{goyal2018graph}.
One popular modeling method uses kernel functions.
The kernel has been theoretically justified; 
for example, 
dot product kernel is shown to be linked to the normalized graph cut via weighted kernel $k$-means~\cite{dhillon2004kernel} and Gaussian kernel is justified via heat kernel~\cite{belkin2003laplacian}. 

Hypergraphs generalize graphs~\cite{Berge84}, and hence are suitable to model data that have multi-way relationships, such as 
videos~\cite{huang2009video} and cells~\cite{cellhypergraph}. 
For hypergraphs, cut-based spectral clustering has also been established~\cite{ZhouHyper,TotalVariation}.
Therefore, from the discussion on graphs, it is natural to model real-valued data as hypergraphs for clustering. 
By looking at multi-way relationships, we aim to gain better clustering results for general data as well as to model data that essentially involves multi-way relationships, such as the examples above.
However, while heuristic modeling as hypergraphs has been done in several domains~\cite{govindu2005tensor,sun2017hypergraph,yu2018modeling}, we are yet to have a modeling framework that is theoretically connected to hypergraph cut problems.

This paper proposes a hypergraph modeling and its spectral embedding framework for clustering, which we theoretically connect to the established hypergraph cut problems. 
This framework models real-valued data as an even order $m$-uniform hypergraphs, all of whose edges connect $m$ vertices. 
For this purpose, we propose a \textit{biclique kernel}, which formulates multi-way similarity, by exploiting the kernel function's ability to model similarity but in a way where we expand from pairs to multiplets. 
We give a theoretical foundation to biclique kernel; a biclique kernel is equivalent to semi-definite even-order tensor (Thm.~\ref{thm:kernelsemidefinite}).
We show that biclique kernel is theoretically connected to the established hypergraph cut problems proposed by
~\cite{ZhouHyper,saito2018hypergraph,ghoshdastidar2015provable} via two problems, weighted kernel $k$-means and heat kernels.
We provide a spectral clustering algorithm for our formulation, which is faster than existing ones ($O(n^{3})$ vs. $O(n^{m})$, where $n$ is the number of data points). 
This speed-up allows us to model as an arbitrarily higher-order hypergraphs in a reasonable computational time. 
We numerically demonstrate that our algorithm outperforms the existing graph and heuristic modeling methods.
Our empirical study also shows that by increasing order of a hypergraph, the performance is gained until a certain point but slightly drops from there.
To our knowledge, it is first time to obtain the behavior of performance of spectral clustering using higher-order (say, $m$$\geq$ 8) uniform hypergraph. 

Our contributions are as follows; i) We provide a formulation to model real-valued data as an even order $m$-uniform hypergraph. ii) We show that our formulation is theoretically linked to the established hypergraph cuts in two ways, weighted kernel $k$-means and heat kernel. iii) We provide a fast spectral clustering algorithm. iv) We numerically show that our method outperforms the standard graph ones and existing heuristic modeling ones. 

This paper is an extended version of our AAAI-22 paper in~\cite{saito2022hypergraph}. 
\textit{All proofs are in Appendix}.

\section{Related Work}
\label{sec:relatedwork}
This section reviews the related work of graph and hypergraph modeling.
There are several approaches for justification of graph modeling via kernel function.
Existing work shows the theoretical connection to the graph cut from the weighted kernel $k$-means~\cite{dhillon2004kernel}, energy minimization problem via continuous heat kernel~\cite{belkin2003laplacian}, and kernel PCA~\cite{bengio2004learning}.
Our approach follows the first two.
A study on hypergraph cut has three approaches. 
One way is a graph reduction way~\cite{Agarwal06}, which also works for non-uniform hypergraphs. 
There are three variants of this; star~\cite{ZhouHyper}, clique~\cite{rod,saito2018hypergraph}, and inhomogeneous~\cite{li2017inhomogoenous,veldt2020hypergraph,liu2021strongly}. 
Other ways are total variation~\cite{TotalVariation,li2018submodular} and tensor modeling for uniform hypergraph~\cite{hu2012algebraic,chen2017fiedler,chang2020hypergraph,ghoshdastidar2014consistency,ghoshdastidar2015provable}.
Our approach follows star and clique ways as well as tensor and its graph reduction approach of~\cite{ghoshdastidar2015provable}. 
We also connect ours to the inhomogeneous way in Appendix.

Comparing to the production of hypergraph cut objectives as above, ways of modeling as hypergraphs have received less attention.
There are various studies to model real-valued data as hypergraphs by heuristic ways~\cite{govindu2005tensor,sun2017hypergraph,yu2018modeling}. 
However, to our knowledge, no studies developed a hypergraph cut-based framework to model real-valued data as hypergraphs.

We remark that, for hypergraph connection,~\citet{whang2020mega} considers weighted kernel $k$-means, but they consider a naive connection between reduced contracted graphs and the standard kernel. Also,~\citet{louis2015hypergraph} and~\citet{ikeda2018finding} consider discrete heat equation, which is connected to random walk. However, those three are different to ours since they do not intend to formulate multi-way relationships.

\section{Tensors and Uniform Hypergraphs}
\label{sec:tensorandhypergraph}
This section introduces notations of tensors and hypergraphs. 
We define an $m$-order tensor as 
$\mathcal{A}$
$\in$
$\mathbb{R}^{n_{1} \times ... \times n_{m}}$, 
whose $(i_{1},i_{2},\ldots,i_{m})$-th element is $a_{i_{1}i_{2}\ldots i_{m}}$$\in$$\mathbb{R}$.
If all the dimensions of an $m$-order tensor $\mathcal{A}$ are identical, i.e., $n_{1}$$=$$\ldots$$=$$n_{m}$$=$$n$, we call this tensor as \textit{cubical}. 
Letting $\mathfrak{S}_{m}$ be a set of permutations $\sigma$ on $\{1,...,m\}$, an even $m$-order cubical tensor is called as \textit{half-symmetric} if for every elements 
\begin{align}
\notag
&\mathcal{A}_{i_{\sigma(1)}...i_{\sigma(m/2)}i_{m/2+\sigma'(1)}...i_{m/2+\sigma'(m/2)}}=\\
&
\label{eq:halfsymmertric}
\mathcal{A}_{i_{m/2+\sigma'(1)}...i_{m/2+\sigma'(m/2)}i_{\sigma(1)}...i_{\sigma(m/2)}},
\forall\sigma,\sigma'\in\mathfrak{S}_{\frac{m}{2}},
\end{align}
see Appendix for examples.
In the following, we assume a half-symmetric even order cubical tensor. 
We define the \textit{mode-$k$ product} of $\mathcal{A}$$\in$$\mathbb{R}^{n_1 \times \ldots \times n_{m}}$ and a vector $\mathbf{x}$$\in$$\mathbb{R}^{n_k}$ as $\mathcal{A}$$\times_{k}$$\mathbf{x}$$\in$$\mathbb{R}^{n_1 \times \ldots \times n_{k-1} \times 1 \times n_{k+1} \times \ldots \times n_{m}}$, whose element is
\begin{align}
\label{eq:modekproduct}
    (\mathcal{A} \times_{k} \mathbf{x})_{i_1\ldots i_{k-1}1i_{k+1}\ldots i_{m}} := \sum_{i_k=1}^{n_k} \mathcal{A}_{i_1\ldots i_{k}\ldots i_{m}} x_{i_{k}}
\end{align}
We define a \textit{contracted matrix} $A^{(m)}$ for a half-symmetric even $m$-order cubical tensor $\mathcal{A}$ as
\begin{align}
\label{eq:defgrammatrix}
A^{(m)} &:= \mathcal{A} \times_{2} \mathbf{1} \times_{3} \cdots \times_{\frac{m}{2}-1} \mathbf{1} \times_{\frac{m}{2}+1} \mathbf{1} \cdots \times_{m} \mathbf{1} 
\end{align}
Note that $A^{(m)}$ is symmetric. 
For details, see~\cite{lim2005singular,qi2005eigenvalues,de2000multilinear}.

An $m$-uniform hypergraph, a generalized graph, can be represented by an $m$-order cubical tensor. 
A {\it hypergraph} $G$ is a set of $(V, E, \mathbf{w}$), where an element of $V$ is called a {\it vertex}, an element of $E$ is called as an {\it edge}, and $\mathbf{w}$ is a vector $\{w(e)\}_{e \in E}$ where $w$$\colon$$E$$\to$$\mathbb{R}^{+}$ associates each edge with a {\it weight}.
When all the edge contains the same number of vertices, we call \textit{uniform}.
A hypergraph is {\it connected} if there is a path for every pair of vertices. 
If an edge contains the same vertex multiple times, we call that this edge has a \textit{self-loop}.
We define an \textit{adjacency tensor} $\mathcal{A}$ for uniform hypergraph, where we assign the weight of edge $e$$=$$\{i_{1}$$,\ldots,$$i_{m}\}$ to $(i_{1},\ldots,i_{m})$-th element of $m$-order cubical tensor. 
A uniform hypergraph is {\it half-undirected} when its adjacency tensor is half-symmetric.
Note that a uniform hypergraph is half-undirected if undirected.
The following assumes that a hypergraph $G$ is uniform, connected, half-undirected, and has self-loops unless noted.
We define the {\it degree} of a vertex $v$$\in$$V$ as $d_i$$=$$\sum_{ e \in E: i \in e}$$w(e)$, and
define a degree matrix $D_{v}$ whose diagonal elements are the degree of vertices.
Let $W_e$$\in$$\mathbb{R}^{|E|\times|E|}$ be a diagonal matrix, whose diagonal elements are weight of edge $e$. 
Let $H$$\in$$\mathbb{R}^{|V|\times|E_{u}|}$ be an \textit{index matrix}, whose element $h(v,e)$ $=$$\sqrt{\rho_{v,e}}$ if a vertex $v$ is connected to an edge $e$, and 0 otherwise, 
where $\rho_{v,e}$ counts how many times the edge $e$ contains the vertex $v$, 
e.g., if edge is $e$$=$$(v,v,v_{1},v_{2})$ for 4 uniform hypergraph, $\rho_{v,e}$$=$$2$.
Other than this tensor way, there is another way to represent hypergraphs as \textit{adjacency matrix}, which contracts hypergraphs into graphs.
There have been three popular ways for this, star~\cite{ZhouHyper} and two variants of clique methods~\cite{rod,saito2018hypergraph}.
In terms of clustering for half-undirected uniform hypergraph, which is our focus, these three different methods produce the same result (see Appendix).  
This paper uses the star method, which contracts a hypergraph into a graph by forming $A_{s}$$:=$$HW_{e}H^{\top}/m$. 
\section{Formulation of Multi-way Similarity}
\label{sec:generalizedkernel}
This section proposes a formulation of multi-way similarity and discusses its properties.
Looking back at a pairwise similarity, kernel functions are a convenient tool to model a similarity.
However, kernel functions consider pairwise similarities, not multi-way similarities. 
The idea to construct a multi-way similarity framework is that we take the benefits of the kernel framework's modeling ability, but at the same time, we expand to multiplets from pairs.

\subsection{Biclique Kernel and Tensor Semi-definitness}

This section formulates multi-way similarity as a \textit{biclique kernel} and discusses its semi-definite property.
For two sets of $m/2$ variables, $\{\mathbf{x}_{i_{\cdot}}\}$ and $\{\mathbf{t}_{l_{\cdot}}\}$, $\mathbf{x}_{i_{\cdot}},\mathbf{t}_{l_{\cdot}}$$\in$$\mathbf{X}$, $\mathbf{X} \subseteq \mathbb{R}^{d}$,
we formulate even $m$ multi-way similarity function $\kappa^{(m)}(\mathbf{x}_{i_{1}},..,\mathbf{x}_{i_{m/2}}$$,\mathbf{t}_{l_{1}},..,\mathbf{t}_{l_{m/2}})$ : $\mathbf{X}^{m/2} $$\times$$\mathbf{X}^{m/2}$$\rightarrow$$\mathbb{R}$ as
\begin{align}
\label{eq:generalizedkernel}
    &\kappa^{(m)} (\{\mathbf{x}_{i_{\cdot}}\},\{\mathbf{t}_{l_{\cdot}}\}\}) := \sum_{\gamma=1}^{m/2}\sum_{\nu=1}^{m/2} \kappa(\mathbf{x}_{i_{\gamma}},\mathbf{t}_{l_{\nu}} ), 
\end{align}
where $\kappa:$$\mathbf{X}$$\times $$\mathbf{X}$$\rightarrow$$\mathbb{R}$ is a standard kernel.
We call $\kappa$ as a \textit{base kernel}. 
By construction, $\kappa^{(m)}$ is also a kernel.
Therefore, we call $\kappa^{(m)}$ as \textit{biclique kernel}. 
Let $\mathcal{K}$ be a \textit{gram tensor} of $\kappa^{(m)}$, i.e., an $m$-order cubical tensor formed by Eq.~\eqref{eq:generalizedkernel}, whose $(i_1,$$...$$,i_m)$-th element is $\kappa^{(m)}($$\mathbf{x}_{i_{1}},$$...$$,\mathbf{x}_{i_{m}})$. Note that $\mathcal{K}$ is half-symmetric due to the construction of $\kappa^{(m)}$. 
Seeing Eq.~\eqref{eq:generalizedkernel}, we can obtain arbitrary even $m$ order biclique kernel from a standard kernel function $\kappa$.
For example of $m$$=$$4$, the biclique kernel is as
\begin{align}
\notag
    &\kappa^{(4)} (\mathbf{x}_{1},\mathbf{x}_{2},\mathbf{t}_1,\mathbf{t}_{2}) \\
    \notag
&=\kappa(\mathbf{x}_{1},\mathbf{t}_1)+\kappa(\mathbf{x}_{1},\mathbf{t}_2)+\kappa(\mathbf{x}_{2},\mathbf{t}_1) + \kappa(\mathbf{x}_{2},\mathbf{t}_2).\\
\notag
&= \exp(-\gamma \|\mathbf{x}_1 - \mathbf{t}_1\|^2) + \exp(-\gamma\|\mathbf{x}_1 - \mathbf{t}_2\|^2) \\
\notag
    &\ \ \ +\exp(-\gamma\|\mathbf{x}_2 - \mathbf{t}_1\|^2) + \exp(-\gamma\|\mathbf{x}_2 - \mathbf{t}_2\|^2).
\end{align}

The biclique kernels are connected to the semi-definite even order tensors, which serves as a theoretical ground of the biclique kernel.
For the standard kernel, a gram matrix for a kernel function is equivalent to a semi-definite matrix~\cite{shawe2004kernel}. 
This characteristic is one of the theoretical foundations of kernel function.
Here, we establish a generalization of this characteristics for the gram tensor $\mathcal{K}$.
We begin with the definition of semi-definiteness of even-order tensors. 
An even $m$-order cubical tensor $\mathcal{A}$ is \textit{semi-definite} if 
\begin{align}
\mathcal{A}\times_{1}\mathbf{x}...\times_{m}\mathbf{x}\geq0, \forall \mathbf{x} \in \mathbb{R}.
\end{align}
Note that the semi-definiteness can be applied only to even order tensors since no odd-order tensors satisfy this semi-definiteness (see Appendix).
For this semi-definiteness of tensors, the following theorem for a tensor formed by a biclique kernel holds. 
\begin{theorem}
\label{thm:kernelsemidefinite}
Given a function $\kappa^{(m)}$$:$$\mathbf{X}^{m/2}$$\times$$\mathbf{X}^{m/2}$$\to$$\mathbb{R}$ defined by $\kappa^{(m)}$$(\{\mathbf{x}_{i_{\cdot}}\}$$,\{\mathbf{t}_{l_{\cdot}}\}\})$$=$$\sum_{\gamma,\nu}$ $\kappa(\mathbf{x}_{i_{\gamma}},\mathbf{t}_{l_{\nu}} )$, where $\kappa$ is a function $\kappa:$$\mathbf{X}$$\times$$\mathbf{X}$$\to$$\mathbb{R}$, then $\kappa$ can be decomposed as $\kappa(\mathbf{x},\mathbf{z})$$=$ $\langle$$\psi(\mathbf{x}),$$\psi(\mathbf{z})\rangle$ if and only if $\kappa^{(m)}$ is half-symmetric and has the $m$-order tensor semi-definite property.
\end{theorem}
This theorem gives a theoretical foundation of the biclique kernel.
Thm.~\ref{thm:kernelsemidefinite} shows that a half-symmetric even-order semi-definite tensor and a biclique kernel are equivalent, which is similar to the foundations of the standard kernel function.

\subsection{Contraction of Biclique Kernel}
Despite of the nice property of Thm.~\ref{thm:kernelsemidefinite}, tensors are practically hard to work with. 
Many tensor problems of generalized common operations in matrix are NP-hard~\cite{hillar2013most}, such as computing eigenvalues.
This motivates us to explore a practically easy while theoretical guaranteed way to deal with biclique kernel.
This section argues that a contracted matrix of a gram tensor can address this issue.

We consider a contacted matrix $K^{(m)}$ (defined in Eq.~\eqref{eq:defgrammatrix}) of a gram tensor $\mathcal{K}$ of the biclique kernel $\kappa^{(m)}$.
We call this contracted matrix $K^{(m)}$ as a \textit{gram matrix} of $\kappa^{(m)}$.
In the following, we see this gram matrix is more computationally efficient while equivalent to the original biclique kernel.
We first observe the following lemma and corollary by contracting a gram tensor into a gram matrix.

\begin{lemma}
\label{lemma:bicliquekernelmatrixation}
Assume $\kappa(\mathbf{x},\mathbf{z})$$=$$\langle \psi(\mathbf{x}),\psi(\mathbf{z}) \rangle_{\kappa}$ is a base kernel of the biclique kernel $\kappa^{(m)}$. Let $\psi_{i}:=\psi(\mathbf{x}_{i})$, and $\Psi := \sum_{l=1}^n\psi_{l}/n$. The gram matrix $K^{(m)}$ of $\kappa^{(m)}$ is equal to a gram matrix formed by a kernel $\kappa':X \times X \rightarrow \mathbb{R}$ as
\begin{align}
\label{eq:kerneldecomp}
    &\frac{\kappa'(\mathbf{x}_i,\mathbf{x}_j)}{n^{m-2}} :=  
\bigl\langle \psi_{i} + \frac{m-2}{2} \Psi, \psi_{j} + \frac{m-2}{2} \Psi \bigr\rangle_{\kappa}.
\end{align}
\end{lemma}
\begin{corollary}
\label{cor:semidefinite}
The gram matrix $K^{(m)}$ is semi-definite.
\end{corollary}
From this lemma, we observe that $K^{(m)}$ is more computationally efficient than $\mathcal{K}$ for the following reason.
Computing Eq.~\eqref{eq:kerneldecomp}, we can rewrite $K^{(m)}$ by using the gram matrix $K$ of the base kernel $\kappa$  as
\begin{align}
    &\frac{K^{(m)}_{ij}}{n^{m-2}} = 
    \label{eq:matrixgeneralizedkernel}
      K_{ij} + \frac{m-2}{2n}(\delta_{i} + \delta_{j}) +  \frac{(m-2)^2}{4n^{2}}\rho 
\end{align}
where 
$\delta_i$ is the sum of $i$-th row of $K$
and $\rho$ is a sum of all the elements of $K$, i.e., $\rho$$= $$\sum_{i,j}$$K_{ij}$.
Since we can pre-compute $\delta_i$, $\delta_j$ and $\rho$ from $K$ in $O(n^{2})$, the overall computational time for $K^{(m)}$ is $O(n^2)$, whereas 
$O(n^{m})$ if we naively form $K^{(m)}$ from the original tensor and Eq.~\eqref{eq:defgrammatrix}.
Note that if we see $K^{(m)}$ as a graph, its degree matrix is equal to a degree matrix $D_v$ of a hypergraph formed by $\mathcal{K}$.
Using this lemma, we obtain the following proposition about equivalence of $\mathcal{K}$ and $K^{(m)}$.
\begin{proposition}
    \label{prop:tensorandmatrix}
    There exists only one kernel $\kappa'$ from a biclique kernel $\kappa^{(m)}$. 
    Also, we can compose only one biclique kernel $\kappa^{(m)}$ from a kernel $\kappa'$ and even-order $m$.
\end{proposition}
This proposition shows that a biclique kernel $\kappa^{(m)}$ and a set of a kernel function $\kappa'$ and even order $m$ are equivalent.
Therefore, Prop.~\ref{prop:tensorandmatrix} is a theoretical guarantee to use a computationally cheaper gram matrix $K^{(m)}$ instead of a computationally expensive gram tensor $\mathcal{K}$.

\section{Hypergraph Cut and Spectral Clustering}
\label{sec:hypergraphcut}
Similarly to the graph case, we want to ground our formulation of multi-way similarity by biclique kernel on a hypergraph cut theory.
This section discusses uniform hypergraph cut, to which we aim to link our formulation later. 
Here we consider partitioning a hypergraph $G$ into two disjoint vertices sets $V_1$,$V_2$$\subset$$V$, $V_1$$\cap$$V_2$$=$$\varnothing$. 
Since the hypergraph edges contain multiple vertices, a generalization from graph cut to hypergraph cut is not straightforward. 
The line of the research of graph contraction ways~\cite{ZhouHyper,saito2018hypergraph} defines hypergraph cut to penalize by a balance of the number of intersected vertices in edge by a partition.
More formally, following~\cite{ZhouHyper}, a hypergraph cut is defined as
\begin{align}
\label{eq:defhypergraphcut}
    \mathrm{Cut}(V_{1},V_{2}) &:= \sum_{e\in E} w(e) |e \cap V_{1}||e \cap V_{2} |/m
\end{align}
We define the normalized hypergraph cut problem as
\begin{align}
    \notag
    \mathrm{NCut}(V_{1},V_{2}) := \mathrm{Cut}(V_{1},V_{2})\left( \mathrm{vol}^{-1}(V_{1}) + \mathrm{vol}^{{-1}}(V_{2}) \right),
\end{align}
where $\mathrm{vol}(V_{j})$$=$$\sum_{i \in V_{j}}d_i$. 
We extend this to $k$-way normalized hypergraph cut problem as
\begin{align}
\label{eq:kwaypartitioningmuniformhypergraph}
    \mathrm{kNCut} (\{V_i\}_{i=1}^{k}) := \sum_{j=1}^{k} \mathrm{NCut}(V_{j}, V \backslash V_{j}).
\end{align}
We can rewrite the minimization problem of Eq.~\eqref{eq:kwaypartitioningmuniformhypergraph} as
\begin{align}
\notag
    &\min \mathrm{kNCut} (\{V_i\}_{i=1}^{k})\\
    \label{eq:deflaphypereigenproblem}
    &= \min \mathrm{trace} Z^{\top}D_{v}^{-1/2}L_{s} D_{v}^{-1/2} Z \mathrm{\ s.t.} Z^{\top}Z = I\\
    \label{eq:defadjhypereigenproblem}
    &= \max \mathrm{trace} Z^{\top} D_{v}^{-1/2}A_{s} D_{v}^{-1/2}Z \mathrm{\ s.t.} Z^{\top}Z = I,
\end{align}
where $L_{s}$$:=$$D_{v}$$-$$A_{s}$ is a \textit{hypergraph Laplacian} and 
\begin{align}
\notag
Z_{ij}:=
\begin{cases}
\sqrt{d_i/\sum_{l \in V_{j}}d_l} & (i \in V_{j})\\
0 & (\mathrm{otherwise}).
\end{cases}
\end{align}
Eq.~\eqref{eq:deflaphypereigenproblem} and Eq.~\eqref{eq:defadjhypereigenproblem} become eigenproblem if we relax $Z$ into real-values. 
As discussed, there are three types of adjacency matrix for hypergraph; star and two cliques.
We can define similar cuts for the other two~\cite{saito2018hypergraph}.
For uniform hypergraphs, which are our focus, these three cuts would produce the same results.
See Appendix for more discussion.

In the line of tensor modeling of uniform hypergraph research~\cite{ghoshdastidar2015provable,ghoshdastidar2017uniform}, $k$-way partitioning problem is also considered, which we refer as \textit{GD}.
Slightly changing from GD, we form an adjacency matrix $A_{g}$ as a contracted matrix of $\mathcal{A}$, i.e.,
\begin{align}
    \label{eq:contractiongd}
    A_{g} := \mathcal{A} \times_{3} \mathbf{1} \cdots \times_{m} \mathbf{1}.
\end{align}
A change from GD is the ``position'' of mode-$k$ products.
The reason for this change is that we want a contraction of half-undirected hypergraph to be symmetric. 
On the other hand, this change does not affect the result in GD since GD assumes undirected hypergraph and symmetric tensor and hence contraction does not change by the position of mode-$k$ products.
The clustering algorithm of GD is to solve the eigenproblem as
\begin{align}
\label{eq:tensorhypergraphcut}
\max \mathrm{trace} Z^{\top} D_{v}^{-1/2} A_{g} D_{v}^{-1/2} Z, \mathrm{s.t.} Z^{\top}Z = I. 
\end{align}
We here show the connection between these two algorithms through the following proposition.
\begin{proposition}
\label{prop:cutequivalent}
For half-symmetric uniform hypergraphs, Eq.~\eqref{eq:tensorhypergraphcut} and Eq.~\eqref{eq:defadjhypereigenproblem} are equivalent.
\end{proposition}
We call solving these eigenproblems as \textit{spectral clustering}.

\section{Proposed Algorithm}
\label{sec:proposedalgorithm}
\begin{algorithm}[!t]
\begin{algorithmic}
\REQUIRE{Data $\mathbf{X}$, $\kappa$, and $m$}
\STATE{Compute $K$ from the base kernel kernel $\kappa$ from data $X$.}
\STATE{Construct a gram matrix $K^{(m)}$ of the biclique kernel $\kappa^{(m)}$ from $K$ by using Eq.~\eqref{eq:matrixgeneralizedkernel}.} 
\STATE{Compute degree matrix $D_{v}$ from $K^{(m)}$ and obtain top $k$-eigenvectors of $D_{v}^{-1/2}K^{(m)}D_{v}^{-1/2}$.}
\STATE{Conduct $k$-means to the obtained top $k$-eigenvectors}
\ENSURE{The clustering result.}
\end{algorithmic}
 \caption{Spectral clustering for hypergraph modeled by generalized kernel.}
 \label{algo:spectralclustering}
 \end{algorithm}

We propose an algorithm for clustering real-valued data via modeling as an even $m$-uniform hypergraph and using hypergraph cut.
The overall algorithm is shown in Alg.~\ref{algo:spectralclustering}.
The core of our algorithm is that we model real-valued data as a hypergraph by our biclique kernel (Eq.~\eqref{eq:generalizedkernel}) and use hypergraph spectral clustering (Prop.~\ref{prop:cutequivalent}).
To do this efficiently, we firstly compute $K^{(m)}$ using Eq.~\eqref{eq:matrixgeneralizedkernel} (the first and second step of Alg.~\ref{algo:spectralclustering}) and then conduct spectral clustering (the third step).
The fourth step uses a simple $k$-means algorithm for obtained eigenvectors to decide the split points, same as the previous studies~\cite{ZhouHyper,ghoshdastidar2015provable}.
The overall computation time of Alg.~\ref{algo:spectralclustering} is $O(n^{3})$, since it takes $O(n^2)$ to compute $K$ as well as $K^{(m)}$, and takes $O(n^{3})$ to compute eigenvectors, which is equivalent to the standard graph spectral methods.
Alg.~\ref{algo:spectralclustering} is faster than 
spectral algorithms naively using Eq.~\eqref{eq:defadjhypereigenproblem}~\cite{ZhouHyper,saito2018hypergraph} and Eq.~\eqref{eq:tensorhypergraphcut}~\cite{ghoshdastidar2015provable} for a hypergraph formed by $\mathcal{K}$.
Both of these cost $O(n^{m})$ to compute $\mathcal{K}$ and $K^{(m)}$, while ours takes  overall $O(n^{3})$. 
This reduction allows us to model as an arbitrary even $m$-uniform hypergraphs in a reasonable computation time, e.g., for a 20-uniform hypergraph $O(n^{3})$ vs.~$O(n^{20})$.
Therefore, Alg.~\ref{algo:spectralclustering} is as scalable as the standard graph methods in terms of $n$, and more scalable than the existing hypergraph methods in terms of $m$.


The question is, what are theoretical justifications for Alg.~\ref{algo:spectralclustering}? 
At this point, it seems ad-hoc to model real-valued data as a hypergraph via biclique kernel for spectral clustering since we do so without any justifications.
To justify Alg.~\ref{algo:spectralclustering}, next two sections connect Alg.~\ref{algo:spectralclustering} to the weighted kernel $k$-means and explain Alg.~\ref{algo:spectralclustering} with Gaussian-type biclique kernel from a heat kernel view.

\section{Kernel $k$-means and Spectral Clustering}
\label{sec:hypergraphspectralclustering}
The graph cut and the standard kernel have a connection through a trace maximization problem via weight kernel $k$-means~\cite{dhillon2004kernel}. 
This section explores a similar connection between our biclique kernel and the hypergraph cuts.
To do so, we first revisit the connection for the standard case and give an alternative way of connection for \textit{any} kernel, instead of the dot product kernel originally discussed in~\cite{dhillon2004kernel}. 
This way is a kernel function approach instead of an explicit feature map approach done in~\cite{dhillon2004kernel}. 
We generalize this way of the graph case to our biclique kernel setting.
We show that this generalized weighted kernel $k$-means objective for our biclique kernel is equivalent to the established cut in Prop.~\ref{prop:cutequivalent}, which we see as a justification of Alg.~\ref{algo:spectralclustering}.

\subsection{Revisiting Spectral Connection}
\label{sec:revisitingspectral}
This section revisits the claim in~\cite{dhillon2004kernel} that weighted kernel $k$-means and graph cuts are connected.
We here give an alternative way of connection.
This alternative way allows us to handle \textit{any} inner product kernels, while the original in~\cite{dhillon2004kernel} only assumes the dot product kernel.
We define clusters by $\pi_j$, a partitioning of points as $\{\pi_j\}_{j=1}^{k}$, and the weighted kernel $k$-means objective for this as 
\begin{align}
\label{eq:kernelkmeans}
    &J(\{\pi_j\}_{j=1}^{k}) := \sum_{\mathbf{x}_{i} \in \pi_{j},j} w(\mathbf{x}_{i})\| \psi(\mathbf{x}_{i}) -  \mathbf{m}_j\|^2,
\end{align}
where $\mathbf{m}_{j}$ is a weighted mean, which is defined as 
\begin{align}
    \mathbf{m}_{j} := \sum_{\mathbf{x}_i \in \pi_j} \frac{w(\mathbf{x}_i)\psi(\mathbf{x}_i)}{s_{j}},\  s_{j}:=\sum_{\mathbf{x}_i \in \pi_j}w(\mathbf{x}_i),
\end{align}
and $\|$$\cdot$$\|$ is a norm induced by \textit{any} inner product forming a kernel function $\kappa(\mathbf{x},\mathbf{y})$$=$$\langle \psi(\mathbf{x}),\psi(\mathbf{y}) \rangle$.
Let $\kappa_{ij}$$:=$$\kappa(\mathbf{x}_{i}, \mathbf{x}_{j})$, $\psi_{i} $$:=$$\psi(\mathbf{x}_i)$, and $w_{i}$$:=$$w(\mathbf{x}_{i})$.
Using the kernel $\kappa$ and its gram matrix $K$ we can rewrite Eq.~\eqref{eq:kernelkmeans} as 
\begin{align}
\notag
    &J(\{\pi_{j}\}_{j=1}^{k}) 
    = \sum_{i \in \pi_{j}, j} w_{i}(\|\psi_{i}\|^2 -2 \langle \psi_{i}, \mathbf{m}_j \rangle + \| \mathbf{m}_j \|^2)\\
    \notag
    &= \sum_{i \in \pi_{j}, j}
    \Bigl( w_{i}\kappa_{ii} - 2w_{i}\sum_{l \in \pi_{j}} \frac{w_{l}}{s_j}\kappa_{il} + w_{i} \sum_{l,r \in \pi_{j}}\frac{w_{l}w_{r}}{s_j^2}\kappa_{lr}\Bigr)\\
    \label{eq:kernelobj}
    &=\sum_{i \in \pi_{j},j}
    w_{i}\kappa_{ii} -
    \sum_{r,l \in \pi_j,j} \frac{w_{r}w_{l}\kappa_{rl}}{s_j} \\
    \label{eq:kernelobjnorm}
    &= \mathrm{trace}W^{1/2}KW^{1/2} - \mathrm{trace}YW^{1/2}KW^{1/2}Y,
\end{align}
where 
\begin{align}
\label{eq:multipleindicator}
Y_{ij}:=
\begin{cases}
\sqrt{w(\mathbf{x}_i)/s_{j}} & (\mathbf{x}_i\in\pi_j)\\
0 & (\mathrm{otherwise}),
\end{cases}
\end{align}
and $W$ is a diagonal matrix whose diagonal element is $w_{i}$.
To minimize Eq.~\eqref{eq:kernelobjnorm}, we want to maximize the second term because the first term is constant w.r.t. the partitioning variable $Y$.
Since $Y^{\top}$$Y$$=$$I$, maximizing the second term is taking the top $k$ eigenvectors of $W^{1/2}$$K$$W^{1/2}$.
Taking $K$ as a graph and $W$ as inverse of the degree matrix, Eq.~\eqref{eq:kernelobjnorm} becomes the relaxed graph cut problem. 
This gives an alternative way to connect the weighted kernel $k$-means and the graph cut.

\subsection{Spectral Connection for Multi-way Similarity}
\label{sec:spectralconnectionformulti}
This section aims to establish a connection between our formulation of multi-way similarity and the hypergraph cut problem, similarly to the graph one. 
To do so, we first generalize a weighted kernel $k$-means for our biclique kernel.
Looking at Eq.~\eqref{eq:kernelobj}, the objective function of weighted kernel $k$-means uses the kernel function $\kappa$ directly.
Therefore, we consider generalizing by replacing $\kappa$ in Eq.~\eqref{eq:kernelobj} to our biclique kernel.
This discussion leads us to define an objective function for weighted kernel $k$-means for multi-way similarity as follows:
\begin{align}
\notag
    &J'(\{\pi_j\}_{j=1}^{k}) :=
    \sum_{i \in \pi_{j},j}
    \sum_{
    \{i_{\cdot}\}
    \subset \pi_{j}
    }
    w_{i}\kappa^{(m)}( i,i_{\cdot},i,i_{\cdot}) \\ 
    \label{eq:defgeneralizedkernelkmeans}
    &
    -\sum_{i,l \in \pi_j,j}
    \sum_{
    \{i_{\cdot}\}, \{l_{\cdot}\} \subset \pi_{j}
    } \frac{w_{i}w_{l}\kappa^{(m)}(i,i_{\cdot},l,l_{\cdot})}{s_{j}},
\end{align}
where we write $i$ instead of $\mathbf{x}_{i}$, and write $i_{\cdot}$ instead of $\{\mathbf{x}_{i_{\cdot}}\}$, a set of $m/2$$-$$1$ variables. 
Seeing the way we form the gram matrix $K^{(m)}$ of $\kappa^{(m)}$ (Eq.~\eqref{eq:defgrammatrix}), we can rewrite Eq.~\eqref{eq:defgeneralizedkernelkmeans} as
\begin{align}
\notag
    &J'(\{\pi_j\}_{j=1}^{k}) =  
    \sum_{\mathbf{x} \in \pi_{j},j}
    w_{i}K^{(m)}_{ii} -
    \sum_{i,l \in \pi_j,j} \frac{w_{i}w_{j}K^{(m)}_{il}}{s_j} \\
    \label{eq:kernelkmeansoverall}
    &= \mathrm{trace} W^{\frac{1}{2}}K^{(m)}W^{\frac{1}{2}} - \mathrm{trace}YW^{\frac{1}{2}}K^{(m)}W^{\frac{1}{2}}Y,
\end{align}
where $Y$ is defined as Eq.~\eqref{eq:multipleindicator} and $K^{(m)}$ is a gram matrix of biclique kernel $\kappa^{(m)}$.
Similarly to the graph case, Eq.~\eqref{eq:kernelkmeansoverall} can be solved by taking top $k$ eigenvectors of $W^{1/2}K^{(m)}W^{1/2}$.

This discussion draws a connection between hypergraph cut and biclique kernel, and justifies Alg.~\ref{algo:spectralclustering}.
Recall that a gram matrix $K^{(m)}$ is obtained by a contraction of a gram tensor $\mathcal{K}$.
Taking a gram matrix $K^{(m)}$ as a contracted matrix from the adjacency tensor of $m$-uniform hypergraph and $W$$=$$D^{-1}_{v}$, where $D_v$ is its degree matrix, 
Eq.~\eqref{eq:kernelkmeansoverall} is equivalent to the hypergraph cut problem (Prop~\ref{prop:cutequivalent} and Eq.~\eqref{eq:defhypergraphcut}).
Thus, the hypergraph cut problem for a hypergraph formed by $\kappa^{(m)}$ is equivalent to the weighted kernel $k$-means objective function for $\kappa^{(m)}$ (Eq.~\eqref{eq:defgeneralizedkernelkmeans}) with a particular weight.
This discussion justifies Alg.~\ref{algo:spectralclustering}, since Alg.~\ref{algo:spectralclustering} turns out to be equivalent to a generalization of weighted kernel $k$-means for $\kappa^{(m)}$. 
Note that since we form $\mathcal{K}$ by $\kappa^{(m)}$, elements of $\mathcal{K}$ can be negative.
This contradicts the assumption that all the weight of an edge is positive.
However, this can be practically resolved in a way that does not affect topological structures, e.g., by adding the same constant to all the data points.
Finally, we remark that we can rewrite Eq.~\eqref{eq:defgeneralizedkernelkmeans} as an Eq.~\eqref{eq:kernelkmeans}-style objective function. 
Let 
\begin{align}
\psi'_{i}:=n^{\frac{m-2}{2}}\left(\psi_{i}+\frac{m-2}{2}\sum_{l=1}^n\frac{\psi_l}{n}\right).
\end{align}
Observing Eq.~\eqref{eq:kernelkmeansoverall}, we can rewrite Eq.~\eqref{eq:defgeneralizedkernelkmeans} as
\begin{align}
\label{eq:kmeansobjectiveforbicliquekernel}
    J'(\{\pi_j\}_{j=1}^{k})  = \sum_{i \in \pi_j,j} w_{i} \| \psi^{'}_{i} - \mathbf{m}'_{j}\|^2, 
\end{align}
where 
\begin{align}
\notag
\mathbf{m}'_j:=\sum_{i \in \pi_j} \frac{w_{i}\psi'_{i}}{s_{j}}, \ s_{j}:=\sum_{i \in \pi_j} w_{i}
\end{align}

We can more generalize this framework to connect inhomogeneous cut and weighted kernel $k$-means, which we discuss in Appendix. 

\section{Heat Kernels and Spectral Clustering}
\label{sec:heatkernels}

This section establishes a connection between heat kernel and biclique kernel to justify Alg~\ref{algo:spectralclustering}.
In the graph case, for a graph made from a gram matrix of Gaussian kernel formed by randomly generated data,
the cut of this graph can be seen as an analog of the asymptotic case of an energy minimization problem of the single variable heat equation using Gaussian kernel as a heat kernel~\cite{belkin2003laplacian}. 
It is also shown that the graph Laplacian converges to the continuous Laplace operator with infinite number of data points~\cite{belkin2005towards}. 
We formulate a multivariate heat equation, to which we can similarly connect our biclique kernel.
We show that the hypergraph cut problem converges to an asymptotic case of the energy minimization problem of this heat equation using our biclique kernel as heat kernel if the number of data points is infinite. 

We define a discrete Laplacian $L^{(m)}_{t,n}$ for $m/2$ variables $\{\mathbf{x}_{i_{\cdot}}\}$$\in$$\mathbf{X}^{m/2}$, $\mathbf{X}$$\subset$$ \mathbb{R}^{d}$ and a function $f$:$\mathbf{X}^{m/2}$$\to$$\mathbb{R}$ which is ``decomposable'' by a single variable function $f'$ as $f(\{\mathbf{x}_{i_{\cdot}}\})$$=$$\sum_{\mu=1}^{m/2} f'(\mathbf{x}_{i_{\mu}})$, $f'$$:\mathbf{X}$$\to$$\mathbb{R}$ as
\begin{align}
\notag
&L^{(m)}_{t,n} f(\{\mathbf{x}_{i_{\cdot}}\}) 
:= - \sum_{\{ \mathbf{y}_{i_{\cdot}} \}} H_{t}^{(m)}(\{ \mathbf{x}_{i_{\cdot}} \},\{\mathbf{y}_{i_{\cdot}}\})f(\{\mathbf{y}_{i_{\cdot}}\}) \\
\label{eq:laplaciantensor}
&+\sum_{\{\mathbf{y}_{i_{\cdot}}\}} \frac{H_{t}^{(m)}(\{ \mathbf{x}_{i_{\cdot}} \},\{\mathbf{y}_{i_{\cdot}}\})f(\{ \mathbf{x}_{i_{\cdot}} \})}{m/2} 
\end{align}
where 
$H^{(m)}_{t}$ is a biclique kernel formed as
\begin{align}
\notag
    &H_{t}^{(m)}(\{ \mathbf{x}_{i_{\cdot}} \},\{\mathbf{y}_{i_{\cdot}}\}) := \sum_{\gamma, \nu =1}^{m/2,m/2} G_{t}(\mathbf{x}_{i_{\gamma}},\mathbf{y}_{i_{\nu}} ),\\
    \notag
    &\mathrm{where\ }
    G_{t}(\mathbf{x},\mathbf{y}) := \frac{\exp\left(-\|\mathbf{x} - \mathbf{y} \|^2/4t\right)}{(4 \pi t)^{d/2}}.
\end{align}
Note that $G_{t}$ is a Gaussian kernel. 
Note also that the coefficient $m/2$ in Eq.~\eqref{eq:laplaciantensor} comes from approximation of heat equation.
Also, define an energy as 
\begin{align}
S_{2}(H^{(m)}_{t},f):=\sum_{\{\mathbf{x}_{i_{\cdot}}\},\{\mathbf{y}_{i_{\cdot}}\}}L^{(m)}_{t,n} f(\{\mathbf{x}_{i_{\cdot}}\})f(\{\mathbf{y}_{i_{\cdot}}\})    
\end{align}
Minimizing this energy with proper constraints is equivalent to the 2-way hypergraph cut problem for a hypergraph formed by $H^{(m)}_{t}$. See Appendix for the detail of both remarks.

We consider to relate discrete operator $L^{(m)}_{t,n}$ to continuous Laplace operator.
Let us begin with the Laplace operator.
Assume a compact differentiable $d$-dimensional manifold $\mathcal{M}$ isometrically embedded into $\mathbb{R}^{N}$, a set of $m/2$ variables $\{ x_{i}\}_{i=1}^{m/2}$, $x_{i}$$\in$$\mathcal{M}$, abbreviated as $\{ x_{\cdot}\}$, and a measure $\mu$.
Consider a problem to obtain a function $f:$$\mathcal{M}^{m/2}$$\to$$\mathbb{R}$, such that
\begin{align}
\notag
    f = &\argmin S_{2}(f)\ \mathrm{s.t.}\ \ \|f\|^2=1, \\
\label{eq:defenergyminproblem}
    &\mathrm{where}\ S_{2}(f):=\|\nabla f\|^2,  f(\{x_{\cdot}\}):=\sum_{i} f'(x_{i}),
\end{align}
and $f'$ is a single variable function $f':\mathcal{M}\to\mathbb{R}$.
From this formulation, the function $f$ in this problem can be described as ``decomposable'' by $f'$, similarly to Eq.~\eqref{eq:laplaciantensor}.
In physics analogy, we can recognize $S_{2}(f)$ as energy, and the problem as an energy minimization problem.
This problem often appears where we want to know a profile minimizing energy, such as velocity profile in fluid dynamics~\cite{courant1962methods}. 
In machine learning, $\nabla f$ can be seen to measure how close each data point is when we embed data from a manifold to the Euclidean space. 
Then, this problem can be thought to find a suitable mapping $f$ best preserving locality, and hence as a clustering algorithm~\cite{belkin2003laplacian}. 

By using Stokes theorem, $\|\nabla f\|^2= \langle \Delta f,f \rangle$, which rewrites this energy minimization problem as
\begin{align}
    \min (S_{2}(f) = \langle \Delta f,f \rangle)
    \label{eq:stokes}
    \mathrm{\ s.t.\ } \|f\|^2 = 1, \langle f,c1 \rangle=0,
\end{align}
where $c$ is constant. 
Since Laplace operator $\Delta$ is semi-definite and $\|f\|^2$$=$$1$ in constraint, the minimizer of Eq.~\eqref{eq:stokes} is given as an eigenfunction of $\Delta f$. 
The first eigenfunction is a constant function that maps variables $x_i \in \mathcal{M}$ to one point.
To avoid this, we introduce the second constraint since the second eigenfunction is orthogonal to the first, which is constant.

We now formulate a multivariate heat equation to analyze $\Delta f$. 
For even $m$ and $m/2$ variables $x_{i} \in \mathcal{M}\subset \mathbb{R}^{d}$, consider the following heat equation on a manifold $\mathcal{M}^{m/2}$ as
\begin{align}
\notag
    \left(\frac{\partial}{\partial t} + \Delta\right)&U(t,\{x_{\cdot}\}) = 0,\  U(0,\{x_{\cdot}\}) = f(\{x_{\cdot}\}),\\
\label{eq:heatequation}
    &\mathrm{where}\ f(\{x_{i_{\cdot}}\})=\sum_{\mu=1}^{m/2} f'(x_{i_{\mu}}).
\end{align}
and $f$ is ``decomposable'' in the same sense as Eq.~\eqref{eq:laplaciantensor} and Eq.~\eqref{eq:defenergyminproblem}.
Eq.~\eqref{eq:heatequation} governs an $m/2$ variables system, which evolves by $m/2$ variables interacting with each other but the initial conditions $f'$ only depend on one variable.
The solution is given as to satisfy 
\begin{align}
    \notag
    U &= \int H_{t}(\{x_{\cdot}\},\{y_{\cdot}\}) U(0,\{y_{\cdot}\}) d\mu(y_{*})\\
    \label{eq:solutiontoheatequation}
    &\mathrm{where}\  d\mu(y_{*}):=\prod_{i=1}^{m/2} d\mu(y_{i})
\end{align}
and $H_t$ is a \textit{heat kernel}. The $d\mu(y_{*})$ corresponds to the decomposable functions as in Eq.~\eqref{eq:laplaciantensor} and Eq.~\eqref{eq:defenergyminproblem}.
For the heat kernel, a well-known example of heat kernel is Gaussian, which gives a solution to one variable Eq.~\eqref{eq:heatequation} when $\mathcal{M}=\mathbb{R}^n$. 
However, it is difficult to obtain a concrete form of heat kernel for a general manifold.
For details of heat kernel, refer to~\cite{rosenberg1997laplacian}.
Since we can prove that $H_{t}^{(m)}$ is also a heat kernel, there exists a heat equation on manifolds $
\mathcal{M}'$ and $\mathcal{M}''$, where $\mathcal{M}'$$=$$\mathcal{M}^{''m/2}$, whose solution is given as Eq.~\eqref{eq:solutiontoheatequation} using $H_t$$=$$H_t^{(m)}$. 
In the following, we consider the heat equation on this manifold $\mathcal{M'}$.

Using Eq.~\eqref{eq:solutiontoheatequation}, we can relate the energy minimization problem to hypergraph cut and justify Alg.~\ref{algo:spectralclustering}. 
The energy minimization problem Eq.\eqref{eq:stokes} in the Euclidean space can be approximated as
\begin{align}
   &S_{2}(f) 
   =\langle \Delta f, f \rangle 
   \label{eq:connectionspectral}
   \approx \frac{1}{t}S_{2}(H^{(m)}_{t},f),
\end{align}
with proper constraints in Eq.~\eqref{eq:stokes} (see Appendix for details).
As discussed when we defined discrete Laplacian (Eq.~\eqref{eq:laplaciantensor}), 
the fourth term $S_{2}(H^{(m)}_{t},f)$ is equivalent to the 2-way hypergraph cut problem using a hypergraph formed by a biclique kernel $H^{(m)}_{t}$ if properly treating constraints. 
Hence, the energy minimization problem Eq.\eqref{eq:stokes} can be seen as a continuous analog to the hypergraph spectral clustering. 
This discussion supports our biclique kernel with Gaussian kernel and Alg.~\ref{algo:spectralclustering}, since Alg.~\ref{algo:spectralclustering} with the Gaussian-type biclique kernel can be thought as an approximation of energy minimization problem Eq.~\eqref{eq:stokes}.
The key observation is that taking a different $m$ corresponds to taking a different manifold satisfying heat equation Eq.~\eqref{eq:heatequation}. 
This is because the biclique kernel $H^{(m)}$ is a different heat kernel for each $m$, and each heat kernel has a manifold, on which Eq.~\eqref{eq:heatequation} holds.
This key observation gives an intuitive insight; 
choosing better $m$ corresponds to choosing a manifold $\mathcal{M}'$ to which the given data space $\mathbf{X}$ fits better.
We conclude this section by theoretically formulating the above discussion in the following theorem.
\begin{theorem}
\label{thm:lapalcianconverge}
Let $\mathcal{M}'$$=$$\mathcal{M}^{m/2}$ be a manifold, on which Eq.~\eqref{eq:heatequation} satisfies with solutions using $H_{t}^{(m)}$. 
Let the data points $\mathbf{x}_1,\cdots \mathbf{x}_n$ be sampled from a uniform distribution on a manifold $\mathcal{M}$, and $f \in C^{\infty}(\mathcal{M}')$. 
Putting $t_n = n^{-1/(2+\alpha)}$, where $\alpha >0$, there exists a constant $C$ such that
\begin{align}
\notag
    \lim_{n \to \infty} C (nt_{n})^{-1}L_{n,t_n}^{(m)} f(\{\mathbf{x}_{i_{\cdot}}\}) = \Delta f(\{\mathbf{x}_{i_{\cdot}}\}) \text{ in probability.}
\end{align}
\end{theorem}
This theorem theoretically supports the discussion in this section; if we have infinite number of data, Eq.~\eqref{eq:laplaciantensor} converges to the continuous Laplace operator and approximation in Eq.~\eqref{eq:connectionspectral} becomes exact.
Note that this theorem is a multivariate version of the result in~\cite{belkin2005towards}.
\section{Experiment}
\label{sec:experiment}
\begin{table*}[!t]
    \small
    \centering
    \begin{tabular}{c|cccccc}
 Kernel and Method & iris & spine  & Ovarian & Hopkins155 \\\hline\hline
Gaussian ($m$$=$$2$, the standard graph)&0.1027 $\pm$ 0.0033 &  0.3191 $\pm$ 0.0025 &   0.1315 $\pm$ 0.0023 &   0.1600\\
Gaussian Ours ($m$$\geq$$4$) &  \textbf{0.0693} $\pm$ 0.0033 &  \textbf{0.2807} $\pm$ 0.0000& \textbf{0.0841} $\pm$ 0.0000 &  \textbf{0.1112}\\
Gaussian~\cite{ghoshdastidar2015provable} & 0.0737 $\pm$ 0.0318 &  0.3000 $\pm$ 0.0000 & 0.1806 $\pm$ 0.0000 &   0.1465\\
Gaussian~(Affine Subpace) & 0.2267 $\pm$ 0.0000 &  0.2839 $\pm$ 0.0000&  0.1690 $\pm$ 0.0023 &  0.1294\\
Gaussian~($d^{H-2}$~\cite{li2017inhomogoenous}) & 0.2407 $\pm$0.0662 & 0.3195 $\pm$ 0.0078 &  0.3317 $\pm$ 0.0892 &  0.1490\\
\hline
Polynomial ($m$$=$$2$, the standard graph)  & 0.2922 $\pm$ 0.0746 &  0.3183 $\pm$ 0.0295 & 0.2043 $\pm$ 0.0780 &  0.2278\\
Polynomial Ours ($m$$\geq$$4$)  & 0.2719 $\pm$ 0.0383  & 0.3142 $\pm$ 0.0452 &  0.1898 $\pm$ 0.0794 &  0.2258\\
Polynomial~\cite{ghoshdastidar2015provable} & 0.4359 $\pm$ 0.0546 & 0.3219 $\pm$ 0.0050 & 0.2817 $\pm$ 0.1201 &   0.2934\\
Polynomial~\cite{yu2018modeling} & 0.3227 $\pm$ 0.0199 &  0.3828 $\pm$ 0.0754 &  0.4399 $\pm$ 0.0093 & 0.2654\\
    \end{tabular}
    \caption{
    Experimental Results. 
    The standard deviation is from randomness involved in the fourth step of Alg.~\ref{algo:spectralclustering}. 
    Since Hopkins155 is the average performance of 155 datasets, this only shows the average.
    Details are in the main text.
    }
    \label{tab:results}
\end{table*}

\begin{figure*}[!t]
\begin{center}
\subfigure[Iris]{%
\includegraphics[width=.24\hsize,clip]{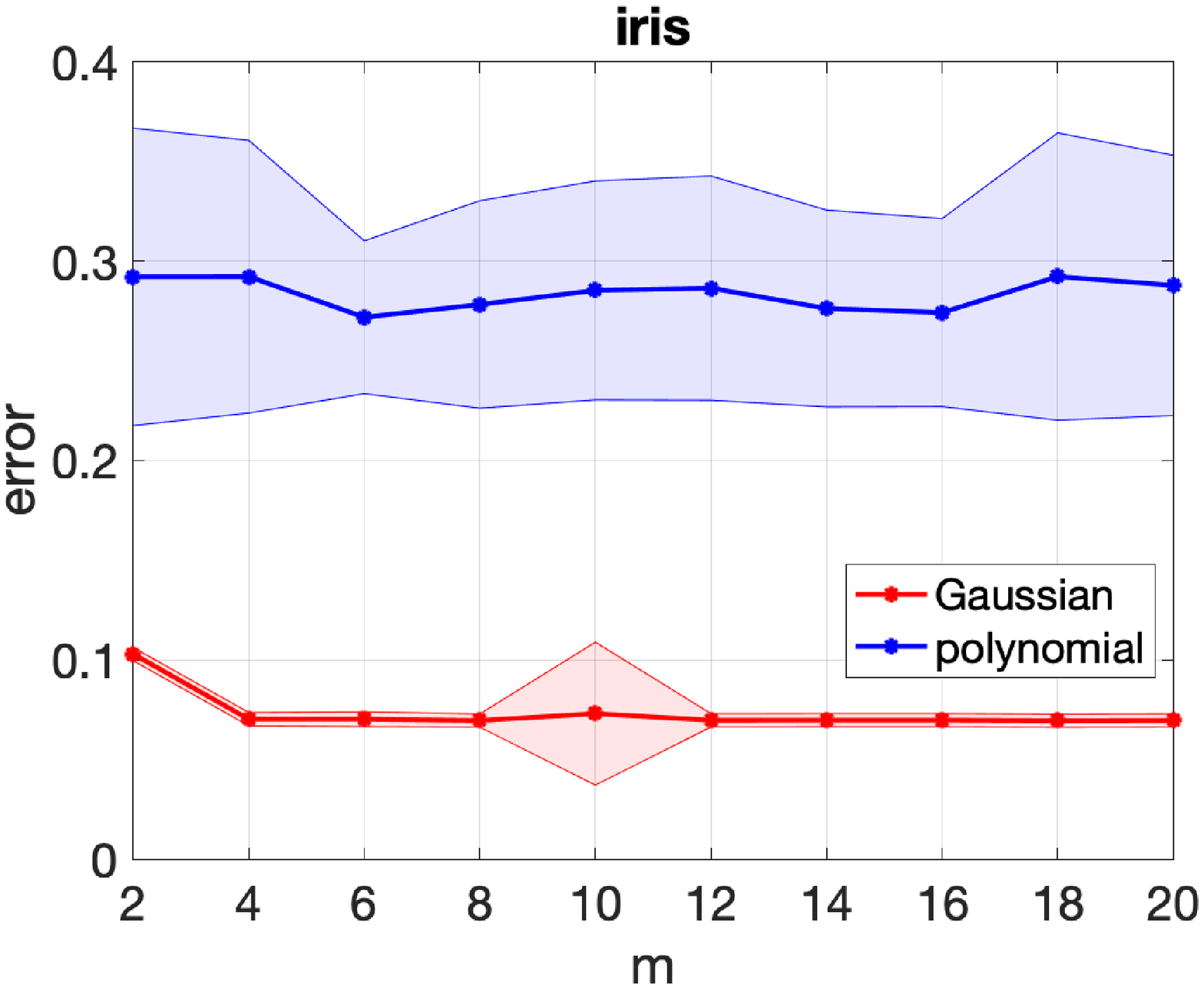}\label{fig:iris}}
\subfigure[Spine]{%
\includegraphics[width=.24\hsize,clip]{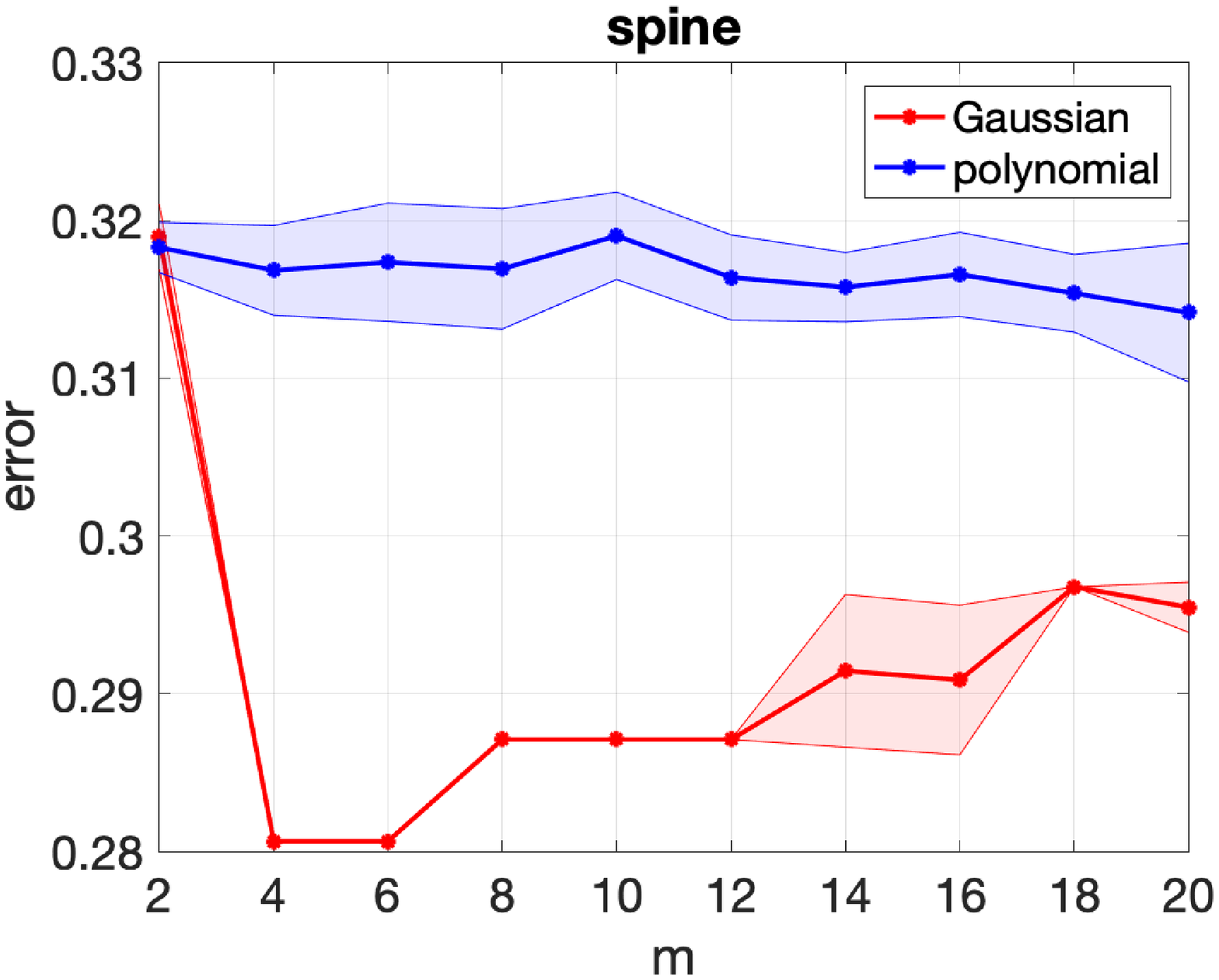}\label{fig:spine}}
~\subfigure[Ovarian]{%
\includegraphics[width=.24\hsize,clip]{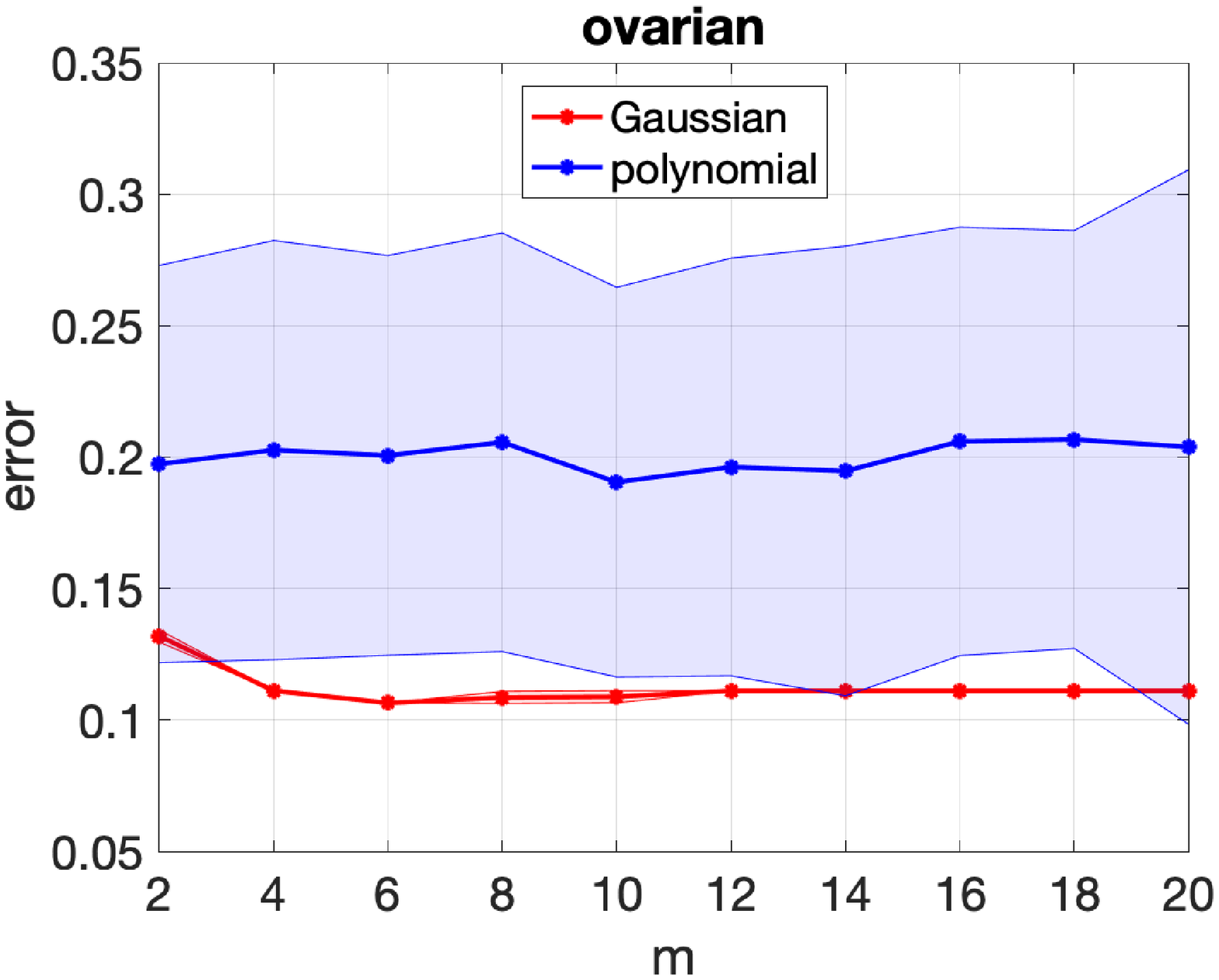}\label{fig:ovarian}}
~\subfigure[Hopkins155]{%
\includegraphics[width=.23\hsize,clip]{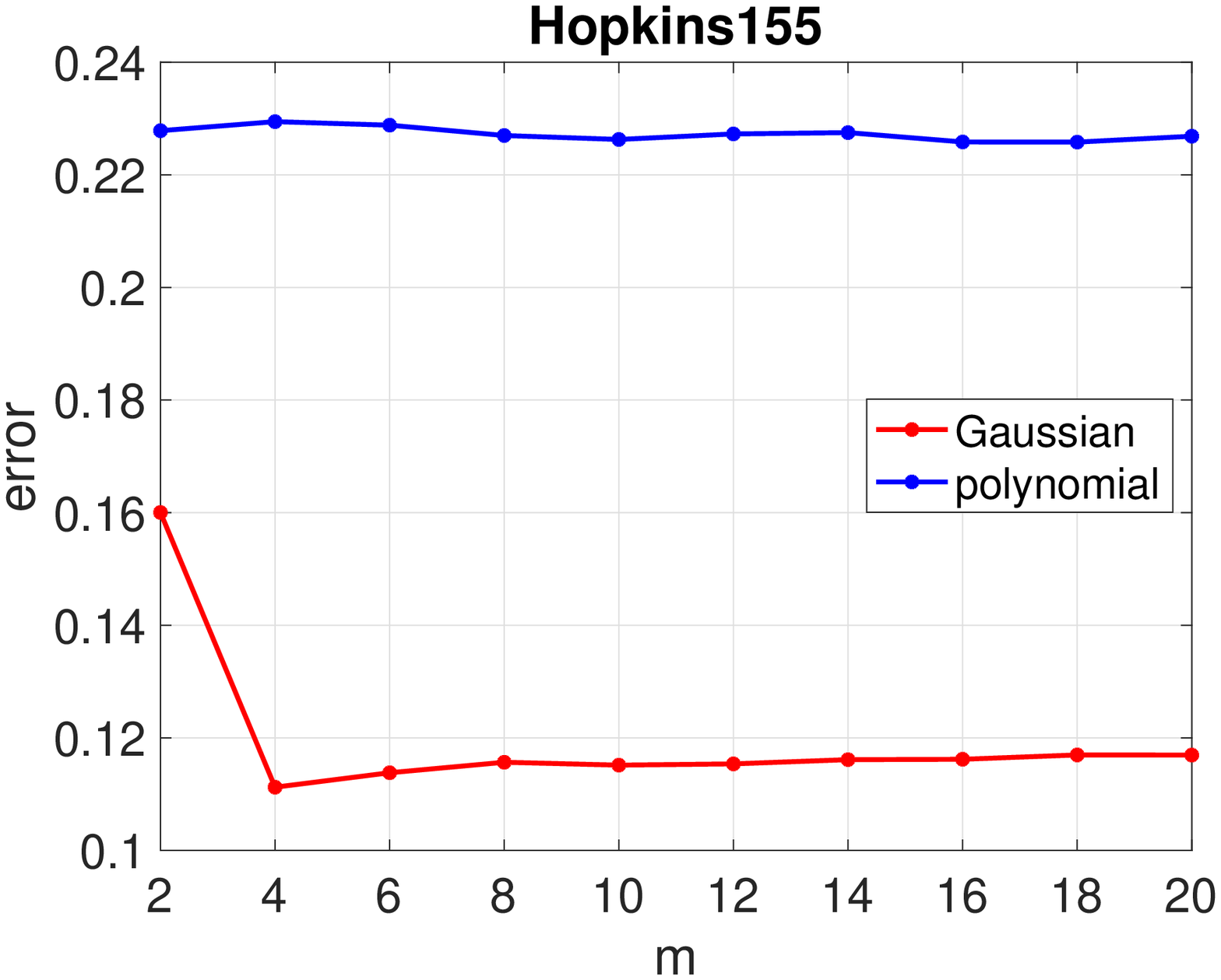}\label{fig:motion}}
\caption{Experimental results. Red shows the result for Gaussian and blue shows for polynomial. The shade shows the standard deviation of the fourth step of Alg.~\ref{algo:spectralclustering}. Since Hopkins155 is the average performance of 155 datasets, this only shows the average.}
\label{fig:exp}
\end{center}
\end{figure*}

This section numerically demonstrates the performance of our Alg.~\ref{algo:spectralclustering} using our formulation of multi-way similarity with biclique kernel.
We evaluated our modeling by comparing the standard kernel and other heuristic hypergraph modelings.
To focus on this purpose, we varied the modelings and kept fixed the cut objective function as Eq.~\eqref{eq:defhypergraphcut}. 
Our experiments were performed on classification datasets, iris and spine from the UCI repository, and ovarian cancer data~\cite{petricoin2002use}.
We also used Hopkins155 dataset~\cite{tron2007benchmark}, which contains 155 motion segmentation datasets.
We used Gaussian kernel ($\kappa(\mathbf{x}_{i},\mathbf{x}_j)$$=$$\exp (- \gamma \| \mathbf{x}_{i}$$-$$\mathbf{x}_{j} \|^2)$) and polynomial kernel ($\kappa(\mathbf{x}_{i},\mathbf{x}_j)$$=$$(\sum_{i}$$x_{i}x_{j}$$+$$ c)^d$) as a base kernel to form a biclique kernel $\kappa^{(m)}$, and conduct Alg.~\ref{algo:spectralclustering}.
We used $m$$=$$2,4$$...,20$.
For comparison, we employed the following types of modeling. 
First, we used $m$$=$$2$, the standard graph method, for both kernels as a baseline. 
Second, we used ad-hoc modeling used in the experiment of~\cite{ghoshdastidar2015provable} for both kernels.
Third, we employed Gaussian-type modeling used in various papers such as~\cite{govindu2005tensor,li2017inhomogoenous}, which is the mean Euclidean distance to the optimal fitted affine subspace. 
Fourth, we used Gaussian-type modeling used in~\cite{li2017inhomogoenous}, which is referred as $d^{H-2}$.
Lastly, for polynomial, we used a generalized dot product form~\cite{yu2018modeling}.
Note that all the hypergraph comparison methods work for any uniform hypergraph.
We can say that we compare five Gaussian-type methods (ours, baseline, \cite{ghoshdastidar2015provable}, affine subspace, and $d^{H-2}$) and four polynomial-type methods (ours, baseline,~\cite{ghoshdastidar2015provable}, and~\cite{yu2018modeling}). 
We restrict hypergraph comparison methods to be $m$$=$$3$ to make the comparison fair in terms of computational time. 
By this, all of the comparisons and ours equally cost $O(n^3)$. 
For the comparisons, we also used the spectral clustering as Eq.~\eqref{eq:defadjhypereigenproblem}, and conduct the forth step of Alg.~\ref{algo:spectralclustering}. 
We used a free parameter $\gamma$$\in$$\{10^{-3}$$,10^{-2}$,...$,10^{5}\}$ for Gaussian, and $d$$\in$$\{1,3,...,9\}$ and $c$$=$$0,1$ for polynomial.
Since the fourth step of Alg.~\ref{algo:spectralclustering} involves randomness at $k$-means, we repeated this step 100 times.
We evaluated our performance on error rate, i.e., (\# of mis-clustered data points)/(\# of data points), same as the previous studies~\cite{ZhouHyper,li2017inhomogoenous}.
We report average errors and standard deviations caused from the fourth step except for Hopkins155. 
Since Hopkins155 contains 155 tasks and standard deviations vary by each task, we only report an average error of 155 tasks, similar to the previous studies~\cite{ghoshdastidar2014consistency,ghoshdastidar2017uniform}. 
Our experimental code is available at github\footnote{\url{https://github.com/ShotaSAITO/HypergraphModeling}}.

We summarize the results in Table~\ref{tab:results} and Fig.~\ref{fig:exp}. 
From Table~\ref{tab:results}, we see that ours with Gaussian kernel outperforms the other methods at all the datasets.
Ours with polynomial kernel also outperforms other polynomial methods.
Additionally, for the most cases in Fig.~\ref{fig:exp}, if we increase $m$, results are improved until a certain point but slightly drop from there. 
This corresponds to the intuition; multi-way relations could be too ``multi'' beyond a certain point: 
Too many relations could work as noise to separate the data.
To our knowledge, it is first time to obtain insights on behaviors of higher-order (say, $m$$\geq$ 8) uniform hypergraph on spectral clustering.
Moreover, for Gaussian methods, the variance for ours is smaller than one for the others.
This means that our methods offer more separated modeling.
Additional results are in Appendix.

\section{Conclusion}
\label{sec:conclusion}
To conclude, we have provided a hypergraph modeling method, and a fast spectral clustering algorithm that is connected to the hypergraph cut problems proposed by~\cite{ZhouHyper,ghoshdastidar2015provable,saito2018hypergraph}.
A future direction would be to explore other constructions of multi-way similarity which can connect to other uniform and non-uniform hypergraph cuts not having kernel characteristics, such as Laplacian tensor ways~\cite{chen2017fiedler,chang2020hypergraph}, total variation and its submodular extension~\cite{TotalVariation,yoshida2019cheeger}. Also, it would be interesting to study more on connections between this work and a general splitting functions of inhomogeneous cut~\cite{li2017inhomogoenous,chodrow2021hypergraph}, e.g., to see which class of splitting functions can be connected to the biclique kernel.
The limitation of our work is that we cannot apply our formulation to an odd-order uniform hypergraph. 
The reason for this limitation is that our biclique kernel is equivalent to half-symmetric semi-definite even-order tensor while odd-order semi-definiteness is indefinite as discussed.


\section{Ethics Statement}
Since this work is in the line of normalized hypergraph cut research, this work shares negative societal impacts with existing works in this line, such as~\cite{ZhouHyper, saito2018hypergraph}.
As pointed out in~\cite{TotalVariation}, the clustering result of normalized cut tends to bias towards large weights. 
This could enhance existing negative biases as a clustering result.

\section{Acknowledgments}
We would like to thank Mark Herbster for valuable discussions.
This research has been made possible thanks to Huawei, who are
supporting Shota Saito's PhD study at UCL.
Shota Saito is also supported by Shigeta Educational Fund.

{
\bibliography{reference}
}

\clearpage

\appendix

\section{Related Work on Graph Spectral Connection}
\label{sec:sepectralconnection}
In addition to the Related Work Section, this section gives a detailed explanation to the connection between the graph cut problem, weighted kernel $k$-means, and heat kernel.
Since our work is a generalization of of the graph case, the connections in the graph case provides a clearer view of our hypergraph case.

For the spectral clustering purpose, the way we model real-valued data into graph has been justified by connecting to graph cut. 
First popular way is modeling by dot product kernel. This way is justified since spectral clustering via this way is equivalent to doing the weighted kernel $k$-means. The reason for this is that the objective function of weighted kernel $k$-means has been shown to be equivalent to the graph cut using the graph formed by the kernel function~\cite{dhillon2004kernel}.
Second and the most popular way is modeling by Gaussian Kernel.
This way is justified because if the number of data is infinite, the spectral clustering via this way converges to the continuous energy minimization problem. In particular, the graph Laplacian converges to the continuous Laplace operator.


\subsection{Graph Cut and Spectral Clustering}
\label{sec:graphcutand}

First, we review the connection between the graph cut and spectral clustering.
We define a graph $G$$=$$(V$$,$$E)$, where $V$ is a set of \textit{vertices} and $E$ is a set of \textit{edges}.
We also define an \textit{adjacency matrix} for a graph $A$, whose $ij$-th element $a_{ij}$ is the weight between $i$-th vertex and $j$-th vertex.
We consider partitioning a graph $G$ into two vertices sets $V_1$,$V_2$$\subset$$V$, $V_1$$\cap$$V_2$$=$$\varnothing$. 
Such partition should happen when a ``dissimilarity'' of two sets is smallest. 
In the graph theory language, such a dissimilarity is called as \textit{cut}; a total weight of edges removed.
We formally define a cut as
\begin{align}
\notag
    \mathrm{Cut}(V_{1},V_{2}) &:= \sum_{i \in V_{1}, j \in V_{2}} a_{ij},
\end{align}
where $a_{ij}$ is a weight of the edge between vertices $i$ and $j$.
Then, we define
\textit{normalized cut} as
\begin{align}
\notag
    \mathrm{NCut}(V_{1},V_{2}) := \mathrm{Cut}(V_{1},V_{2})\left(\frac{1}{\mathrm{vol}(V_{1})} + \frac{1}{\mathrm{vol}(V_{2})}\right),
\end{align}
where $\mathrm{vol}(V)$$:=$$\sum_{i \in V}$$d_{i}$.
Minimizing this can partition the vertex set into two subsets.  
We extend to $k$-way partitioning problem where we partition into $k$ subsets, $V_{i} (i=1,\ldots,k)$, where $V_{i}$$\cap$$V_{j}$$ =$$\varnothing$ if $i\neq j$ and $\cup_{i=1}^{k}$$V_{i}$$=$$V$, formulated as 
\begin{align}
\label{eq:kwaynormalized}
    \min \mathrm{kNCut}(\{V_{i}\}_{i=1}^{k}) := \min \sum_{i=1}^{k} \mathrm{NCut} (V_{i}, V \backslash V_{i}).
\end{align}
We consider to rewrite this problem as a matrix eigenproblem. 
We here introduce an indicator matrix $Z$$\in$$\mathbb{R}^{N \times k}$ as
\begin{align}
\notag
Z_{ij}:=
\begin{cases}
\sqrt{d_i/\sum_{l \in V_{j}}d_l} & (i \in V_{j})\\
0 & (\mathrm{otherwise}).
\end{cases}
\end{align}
We note that $Z^{\top}Z=I$. Using this indicator matrix, this problem can be rewritten as
\begin{align}
\label{eq:grapheigenproblem}
    &\min \mathrm{kNCut}(\{V_{i}\}_{i=1}^{k}) \\
    &=\max_{Z} \mathrm{trace}(Z^{\top} D^{-1/2} A D^{-1/2} Z) \  \mathrm{s.t.} \ Z^{\top} Z = I,\\
\label{eq:graphlaplacianeigenproblem}
    &=\min_{Z} \mathrm{trace}(Z^{\top} D^{-1/2} L D^{-1/2} Z) \  \mathrm{s.t.} \ Z^{\top} Z = I,
\end{align}
where $L$ is a \textit{graph Laplacian} defined as $L:=D-A$.
This discrete optimization problem is known to be NP-hard. 
However, if we relax $Z$ into real value, Eq.~\eqref{eq:grapheigenproblem} becomes an eigenproblem of $D^{-1/2}AD^{-1/2}$, which shows that the graph cut problem can be written as an eigenproblem.
Solving eigenproblem Eq.~\eqref{eq:grapheigenproblem} is called \textit{spectral clustering}.

From the graph cut discussion, the most common modeling algorithm is as follows; i) forming a graph by a kernel function $\kappa$ 
\begin{align}
\notag
    \kappa (\mathbf{x},\mathbf{x}') := \langle\psi(\mathbf{x}),\psi(\mathbf{x}')\rangle.
\end{align}
for the given real-valued data. ii) applying spectral clustering 
Eq.~\eqref{eq:grapheigenproblem}.

While this algorithm ``models'' real-valued data by a kernel function without any justifications,  
next two sections justify this algorithm.
One shows that this algorithm is equivalent to solving the weighted kernel $k$-means problem.
The other shows that with the infinite number of data this algorithm converges to solving the energy minimization problem.

\subsection{Weighted Kernel $k$-means and Spectral Clustering.}
\label{sec:weightedkernel}

This section reviews the connection between spectral clustering and the weighted kernel $k$-means. 
While these share the same purpose in terms of clustering, the formulations seem quite different.
However, the formulations of these are linked via the same trace maximization problem.
This discussion leads to model real-valued data into graphs by considering kernels for clustering.
Taking a certain weight for the weighted kernel $k$-means, those are equivalent through a trace maximization problem~\cite{dhillon2004kernel}.

The weighted kernel $k$-means algorithm is to partition data points in feature space by minimizing the sum of the square of the distance between each data point belonging to the cluster and its centroid.
We define clusters by $\pi_j$, and a partitioning of points as $\{\pi_j\}_{j=1}^{k}$.
We denote the ``energy'' of a cluster $\pi_j$  is as
\begin{align}
\notag
    J(\pi_j) := \sum_{\mathbf{x}_{i} \in \pi_{j}} w(\mathbf{x}_{i})\| \psi(\mathbf{x}_{i}) -  \mathbf{m}_j\|^2_{2},
\end{align}
where $\|\cdot\|_{2}$ is a norm induced from the dot product\footnote{The main text shows that the discussion in this section can be extended from a dot product kernel to \textit{any} kernel.}, $w(\mathbf{x}_i)$ is a weight at $\mathbf{x_i}$ and $\mathbf{m}_j$ serves as a weighted mean of the cluster, which can be written as
\begin{align}
\notag
    \mathbf{m}_{j} := \sum_{\mathbf{x}_i \in \pi_j} \frac{w(\mathbf{x}_i)\psi(\mathbf{x}_i)}{s_{j}}, s_{j}:=\sum_{\mathbf{x}_i \in \pi_j}w(\mathbf{x}_i).
\end{align}
By taking $J(\{\pi_{j}\}_{j=1}^{k})$$:=$$\sum_{j=1}^{k}$$J(\pi_j)$,
this can be the $k$-means objective function and is often minimized by the EM-type algorithm~\cite{bishop2007}.
We can rewrite $J(\{\pi_{j}\}_{j=1}^{k})$ as a trace maximization problem;
\begin{align}
\label{eq:kmeansobj}
&J(\{\pi_{j}\}_{j=1}^{k}) =\mathrm{trace}W^{1/2} K W^{1/2} - \mathrm{trace}Y^{\top} W^{1/2} K W^{1/2} Y,
\end{align}
where $W$ is a diagonal matrix whose $i$-th element is $w(\mathbf{x}_i)$ and $K$ is a gram matrix formed by the dot product kernel, and indicator matrix $Y$$\in$$\mathbb{R}^{n\times k}$
\begin{align}
\notag
Y_{ij}:=
\begin{cases}
\sqrt{w(\mathbf{x}_i)/\sum_{\mathbf{x}_l \in \pi_j}w(\mathbf{x}_l}) & (\mathbf{x}_i \in \pi_j)\\
0 & (\mathbf{x}_{i} \notin \pi_j)
\end{cases}
\end{align}
We note that $YY^{\top}$$=$$I$.
Since the first term of Eq.~\eqref{eq:kmeansobj} is fixed with respect to the variable $Y$, we want to maximize the second term of Eq.~\eqref{eq:kmeansobj} in order to minimize the $k$-means objective function. 
Exploiting the linear algebra theory, maximizing $\mathrm{trace}(Y^{\top} W^{1/2} K W^{1/2} Y)$ can be realized by taking $Y$ as the largest $k$ eigenvectors of $W^{1/2} K W^{1/2}$. 
When we take $W$$=$$D^{-1}$ and $K$$=$$A$, the objective function Eq.~\eqref{eq:kmeansobj} is equivalent to the normalized graph cut objective function Eq.~\eqref{eq:grapheigenproblem}.
Now we see the connection between spectral clustering and weighted kernel $k$-means.

Above all, modeling into graphs by the dot product kernel is justified; if we conduct spectral clustering to a graph formed by the dot product kernel, this is equivalent to the weighted kernel $k$-means with a particular weight.
For more details, see~\cite{dhillon2004kernel,tuto}.

\subsection{Heat Kernel and Spectral Clustering}
\label{sec:heatkerneland}
This section reviews connection between spectral clustering and the heat kernel.
Heat kernel is closely related to the energy minimization problem using Laplace operator, while the graph cut also can be seen the energy minimization problem using graph Laplacian.
This section explains that these two continuous and discrete problem is connected.
More specifically, for a graph made from a gram matrix of Gaussian kernel formed by randomly generated data,
the cut of this graph can be seen as an analog of the asymptotic case of an energy minimization problem of the heat equation using Gaussian kernel as a heat kernel~\cite{belkin2003laplacian}. 
It is also shown that the graph Laplacian converges to the continuous Laplace operator with infinite number of data points~\cite{belkin2005towards}. 

First, we define the energy minimization problem. 
Assume a compact differentiable $d$-dimensional manifold $\mathcal{M}$ isometrically embedded into $\mathbb{R}^{N}$, a variable $x$$\in$$\mathcal{M}$, and a measure $\mu$.
We consider a problem to obtain a function $f:$$\mathcal{M}^{m/2}$$\to$$\mathbb{R}$, that minimizes the energy as
\begin{align}
\label{eq:energyminimization}
S_{2}(f):=\|\nabla f\|^2 \mathrm{\ s.t.\ } \|f\|^2=1.
\end{align}

We can rewrite Eq.~\eqref{eq:energyminimization} using Laplace operator as
\begin{align}
\label{eq:energyminimizationlaplacian}
    \|\nabla f\|^2 = \langle \nabla f, \nabla f \rangle = \langle \Delta f, f \rangle \mathrm{\ s.t.\ } \|f\|^2=1, \langle f, c\mathbf{1} \rangle,
\end{align}
where $c$ is constant. 
The second equality follows from the Stokes theorem. 
We introduce the additional constrains $\langle f, c\mathbf{1} \rangle$ in order to avoid the trivial solution to this problem, which is $f=c1$.
The reason why this works as a constraint comes from a nature of semi-definiteness of Laplace operator $\Delta$. See more discussions in the main text and Sec.2 in~\cite{belkin2003laplacian}.

We now discuss heat equation and heat kernel, which are useful tools to analyze $\Delta f$.
We pay attention to the term $\Delta f$ because this term is the main ``actor'' of the energy minimization problem in Eq.~\eqref{eq:energyminimizationlaplacian}.
Consider a variable $x \in \mathcal{M}$. The heat equation on $\mathcal{M}$ is as
\begin{align}
\label{eq:heatequationsinglevariable}
    \left(\frac{\partial}{\partial t} + \Delta\right)U(t,x) = 0,\ \  U(0,x) = f(x)
\end{align}
The solution is given as to satisfy 
\begin{align}
    \label{eq:solutiontoheatequationsingle}
    U = \int H_{t}(x,y) U(0,y) d\mu(y)
\end{align}
where $H_t$ is a \textit{heat kernel}.
A well-known example of heat kernel is Gaussian kernel as
\begin{align}
\notag
    G_{t}(x,y) = \frac{1}{(4 \pi t)^{d/2}} \exp \left( - \frac{\|x - y\|^2}{4t} \right),
\end{align}
which gives a solution to one variable Eq.~\eqref{eq:heatequation} when $\mathcal{M}=\mathbb{R}^n$. 
However, it is difficult to obtain a concrete form of heat kernel for a general manifold.

We give a solution in asymptotic case when $t \to 0$.
Locally, we can approximate $H_{t}(x,y) = G_{t}(x,y)$ when $t$ and $\|x-y\|$ are small~\cite{rosenberg1997laplacian,belkin2003laplacian}. 
Together with $\lim_{t \to 0}\int_{\mathcal{M}} d\mu(y) G_{t}(x,y)f(y) = f(x)$ and $\lim_{t \to 0}\int_{\mathcal{M}} d\mu(y) G_{t}(x,y) = 1$, for small $t$ and discrete values $ \mathbf{x}_{1},\ldots,\mathbf{x}_{n}$ instead of continuous value we can approximate as
\begin{align}
\label{eq:laplacian}
    \Delta f(\mathbf{x}_{i}) \approx \sum_{j} G_{t}(\mathbf{x}_{i},\mathbf{x}_{j}) f(\mathbf{x}_{i}) - \sum_{j} G_{t}(\mathbf{x}_{i},\mathbf{x}_{j}) f(\mathbf{x}_{j}).
\end{align}
The right hand side of Eq.~\eqref{eq:laplacian} is equal to the graph Laplacian $L$ for a graph whose adjacency matrix is a gram matrix for the Gaussian Kernel.
Following Eq.~\eqref{eq:laplacian}, we can relate the original energy minimization problem and graph cut problem (Eq.~\eqref{eq:graphlaplacianeigenproblem}) as
\begin{align}
\notag
    \| \Delta f \|^2 \approx Y^{\top} L Y
\end{align}
with proper constrains. 
By properly introducing ``normalizing'' constrains, the continuous energy minimization problem is corresponds to 2-way normalized cut problem.

This discussion can justify the modeling by a kernel function for spectral clustering.
The reason is that the graph cut problem for a graph made from a Gaussian kernel can be seen as a approximated continuous energy problem of an asymptotic case of the heat equation.
Therefore, modeling by kernel for spectral clustering is a discrete analog of energy minimization problem.

Finally, we remark that the approximation becomes exact when a number of randomly generated data is infinite (See Thm.3.1 in~\cite{belkin2005towards}). On more discussion of this, we refer to~\cite{belkin2003laplacian,belkin2005towards}.

\subsection{From Graph to Hypergraph}

\begin{table*}[!t]
\centering
    \small
    \caption{List of objective functions of $m$-uniform hypergraph spectral connection and the corresponding pairwise ones. 
    If we model by the kernels as listed, the $k$-way cut, the weighted kernel $k$-means with a particular weight, energy minimization problem using Laplace operator are equivalent to the spectral clustering. Details are discussed in the main text.}
    \label{tab:comparison}
    \begin{tabular}{c|cc}
        & Graph (pairwise) & Hypergraph (multi-way)\\
        \midrule
        Kernel & $ \displaystyle \kappa(\psi(\mathbf{x}_{1}), \psi(\mathbf{x}_{2})) := \langle \psi(\mathbf{x}_{1}), \psi(\mathbf{x}_{2}) \rangle$  & $\displaystyle \kappa^{(m)}(\{\mathbf{x}_{i_{\cdot}}\},\{\mathbf{t}_{i_{\cdot}}\}) := \sum_{\gamma,\nu} \kappa(\mathbf{x}_{i_{\gamma}},\mathbf{t}_{j_{\nu}} ) $ \\
        \midrule
        $k$-way cut & $ \displaystyle \sum_{j=1}^{k}\sum_{i_{1} \in V_{j}, i_{2} \in V\backslash V_{j}} w_{i_1 i_2}$  &  $\displaystyle \sum_{i=1}^{k} \sum_{e \in E}\sum_{j_{1},j_{2} \in e; j_{1} \in V_{j}, j_{2} \in V_{i} \backslash V} w(e)$. \\
        Spectral clustering & \begin{tabular}{c}
             The top $k$ largest eigenvectors of \\graph adjacency matrix $A$ \\
             / gram matrix $K$ 
        \end{tabular}& \begin{tabular}{c}
              Top $k$ largest eigenvectors of \\ hypergraph adjacency matrix $A$ /\\ gram matrix $K^{(m)}$. 
        \end{tabular}\\
        Kernel $k$-means& $ \displaystyle \sum_{j=1}^{k}\sum_{\mathbf{x}_{i} \in \pi_{j}} w(\mathbf{x}_{i})\| \psi(\mathbf{x}_{i}) -  \mathbf{m}_j\|^2$  & $ \displaystyle \sum_{\mathbf{x}_{i} \in \pi_{j}} w'(\mathbf{x}_{i}) \| \psi'(\mathbf{x}_{i}) -  \mathbf{m}'_j \|^2$\\
        Heat Kernel & \begin{tabular}{c}
        $\langle\Delta f, f\rangle$, s.t., $\langle f,c1 \rangle$ where $f$ obeys\\
        $\displaystyle \left( \frac{\partial}{\partial t} + \Delta \right)U(t,x)=0$,\\ $U(0,x)=f(x)$
        \end{tabular}  &
        \begin{tabular}{c}
        $\langle\Delta f, f\rangle$, s.t., $\langle f,c1 \rangle$ where $f$ obeys\\
        $\displaystyle \left( \frac{\partial}{\partial t} + \Delta \right)U(t,\{x_{\cdot}\})=0$,\\ $\displaystyle U(t,\{x_{\cdot}\})=f(\{x_{\cdot}\}) = \sum_{i}^{m/2} f'(x_i)$      
        \end{tabular} 
    \end{tabular}
\end{table*}
This section relates to the discussion of graph here to the discussion in the main text.
Throughout this section, we justify the graph modeling algorithm.
Similarly to this, in the latter of the main text, we justify our biclique kernel by generalizing spectral connection in the graph.
Our biclique kernel generalizes from the discussions on a graph and a kernel to the discussions on a biclique kernel and hypergraph.
We summarize the relationship of spectral connection in graph and in hypergraph (this work) in Table~\ref{tab:comparison}.

\section{Examples of Half-Symmetric Tensors}
\label{sec:examplesof}
This section discusses examples of half-symmetric tensors defined as Eq.~\eqref{eq:halfsymmertric}.
As an example, we consider 4-order half-symmetric cubical tensor $\mathcal{A}$.
Let us think about the element (1,2,3,4) of half-symmetric tensor $\mathcal{A}$. 
Then
\begin{align}
\notag
    &\mathcal{A}_{1234} = \mathcal{A}_{2134} = \mathcal{A}_{1243} = \mathcal{A}_{2143} = \\
    \notag
    &\mathcal{A}_{3412} = \mathcal{A}_{4312} = \mathcal{A}_{3421} = \mathcal{A}_{4321}.
\end{align}

More general, for elements $(i_{1},i_{2},i_{3},i_{4})$ where $i_{j} \neq i_{l}$ if $j \neq l$, the tensor $\mathcal{A}$ is
\begin{align}
\notag
    &\mathcal{A}_{i_{1}i_{2}i_{3}i_{4}} = \mathcal{A}_{i_{2}i_{1}i_{3}i_{4}} = \mathcal{A}_{i_{1}i_{2}i_{4}i_{3}} = \mathcal{A}_{i_{2}i_{1}i_{4}i_{3}} = \\
    \notag
    &\mathcal{A}_{i_{3}i_{4}i_{1}i_{2}} = \mathcal{A}_{i_{4}i_{3}i_{1}i_{2}} = \mathcal{A}_{i_{3}i_{4}i_{2}i_{1}} = \mathcal{A}_{i_{4}i_{3}i_{2}i_{1}}.
\end{align}

Then the natural question to ask is what happens if the elements contains the same index. 
Let us think about the element (1,1,2,3) of half-symmetric tensor $\mathcal{A}$. 
Then
\begin{align}
\notag
    &\mathcal{A}_{1123} = \mathcal{A}_{1132} = \mathcal{A}_{2311} = \mathcal{A}_{3211}.
\end{align}
However, if we think about the elements contains the same index, but in a different ``half'', e.g., (1,2,1,3), then
\begin{align}
\notag
    &\mathcal{A}_{1213} = \mathcal{A}_{2113} = \mathcal{A}_{1231} = \mathcal{A}_{2131}\\
    \notag
    &\mathcal{A}_{1312} = \mathcal{A}_{3112} = \mathcal{A}_{1321} = \mathcal{A}_{3121}.
\end{align}

Needless to say, we do not have such a permutation of index for the ``diagonal'' elements, where all the indices are the same number, e.g., ($1,1,1,1$) and $\mathcal{A}_{1111}$.

\section{On Tensor Semi-definiteness and Proof for Theorem~\ref{thm:kernelsemidefinite}}
\label{sec:proofforthmkernel}
This section gives a detail for tensor semi-definiteness and full proofs for Thm.~\ref{thm:kernelsemidefinite}.
Note that our definition of semi-definiteness is not our own and follows the existing work such as~\cite{qi2005eigenvalues,hu2012algebraic,hillar2013most}.

\subsection{On Tensor Semi-definiteness}
We start with reviewing definition of semi-definiteness of an even $m$-order tensor. 
An even $m$-order tensor is semi-definite when $\forall\mathbf{x}\in \mathbf{R}$,
\begin{align}
    \label{eq:defsemidefinite}
    \mathcal{A} \times_{1} \mathbf{x} \ldots \times_{m}\mathbf{x}=\sum_{i_{1}\ldots i_{m}} \mathcal{A}_{i_{1}\ldots i_{m}} \mathbf{x}_{i_{1}}\ldots \mathbf{x}_{i_{m}} \geq 0.
\end{align}
Let us assume a tensor $\mathcal{A}$ is 3-order cubical tensor. 
From this definition, polynomial for $\mathbf{x}$ formed from odd order tensor can take both positive and negative values, such as
\begin{align}
\notag
    \mathcal{A} \times_{1} -\mathbf{x} \times_{2}  - \mathbf{x} \times_{3} - \mathbf{x}= - \mathcal{A} \times_{1} \mathbf{x} \times_{2} \mathbf{x} \times_{3} \mathbf{x}.
\end{align}
This means that, the polynomial Eq.~\eqref{eq:defsemidefinite} for $\mathbf{x}$ and for $-\mathbf{x}$ take different signs.
However, the semi-definiteness requires the polynomial Eq.~\eqref{eq:defsemidefinite} to be positive for all vectors, including $\mathbf{x}$ and $-\mathbf{x}$.
Therefore, there is no odd-order semi-definite tensors for this definition.
For more discussion on tensor semi-definiteness, see Sec.~11 in~\cite{hillar2013most} and~\cite{qi2005eigenvalues}. 

\subsection{Proof for Thm.~\ref{thm:kernelsemidefinite}}
We start with the `only if' direction.
Following the definition of semi-definiteness of tensors,
\begin{align}
\notag
    &\mathcal{K} \times_{1} \mathbf{v} \ldots \times_{m}\mathbf{v}\\
    \notag
    &=\sum_{i_{1}\ldots i_{m}=1}^{n}
    v_{i_{1}}\ldots v_{i_{m}}
    \kappa^{(m)}(\{x_{i_{\mu}}\}_{\mu=1}^{m/2},\{x_{i_{\mu}}\}_{\gamma=m/2+1}^{m}) \\
    \notag
    &= \sum_{i_{1}\ldots i_{m}}\sum_{j_{1}, j_{2}=1}^{m/2} v_{i_{1}}\ldots v_{i_{m/2}}\kappa(\mathbf{x}_{i_{j_{1}}},\mathbf{x}_{i_{i_{m/2+j_{2}}}})v_{i_{m/2+1}}\ldots v_{i_{m}} \\
    \notag
    &= \left\langle \sum_{i_{1}\ldots i_{m/2},j_1 } v_{i_{1}}\ldots v_{i_{m/2}}\psi(\mathbf{x}_{i{j_1}}), \right. \\
    \notag
    &\left.
    \sum_{i_{m/2+1}\ldots i_{m}, i_{j_2}} v_{i_{m/2+1}}\ldots v_{i_{m}}\psi(\mathbf{x}_{i_{m/2+j_2}}) \right\rangle\\
    \notag
    &= \left\| \sum_{i_{1}\ldots i_{m/2},j_1 } v_{i_{1}}\ldots v_{i_{m/2}}\psi(\mathbf{x}_{i_{j_{1}}}) \right\|^2 \geq 0.
\end{align}
This shows that a gram tensor $\mathcal{K}$ is semi-definite.

We now move to prove `if' direction.
We construct the space such as
\begin{align}
\notag
    \mathscr{F} = \Biggl\{& \sum_{i_1,\ldots,i_{m/2}=1}^{l}\alpha_{i_{1}}\alpha_{i_{2}}\ldots\alpha_{i_{m/2}}\kappa(\{\mathbf{x}_i\}_{i=1}^{m/2},\cdot) \Biggr.\\
    \notag
    &\Biggl. | l \in \mathbb{N},\mathbf{x}_{i} \in X, \alpha_{i_{\cdot}} \in \mathbf{R},i=1,\ldots,l\Biggr\}.
\end{align}

We emphasize that the element of the set $\mathscr{F}$ is a function that takes $m/2$ arguments. 
Note that we have used $\cdot$ to indicate the position of the argument of the function.
Let the function $f,g \in \mathscr{F}$ as
\begin{align}
\notag
    f(\{\mathbf{x}_i\}_{i=1}^{m/2}) = \sum_{i_{\cdot}=1}^{l}\alpha_{i_{1}}\alpha_{i_{2}}\ldots\alpha_{i_{m/2}}\kappa(\{\mathbf{t}_i\}_{i=1}^{m/2},\{\mathbf{x}_i\}_{i=1}^{m/2})\\
\notag
    g(\{\mathbf{x}_i\}_{i=1}^{m/2}) = \sum_{l_{\cdot}=1}^{n}\beta_{l_{1}}\beta_{l_{2}}\ldots\beta_{l_{m/2}}\kappa(\{\mathbf{z}_i\}_{i=1}^{m/2},\{\mathbf{x}_i\}_{i=1}^{m/2})
\end{align}
We now introduce inner product $\langle f,g \rangle$ as follows;
\begin{align}
\label{eq:innerproduct}
    &\langle f,g \rangle \\
    \notag
    &:= \sum_{i_{\cdot}=1}^{l}\sum_{j_{\cdot}=1}^{n} \alpha_{i_{1}} \alpha_{i_{2}}\ldots\alpha_{i_{m/2}} \beta_{l_{1}}\beta_{l_{2}}\ldots\beta_{l_{m/2}} \kappa (\{\mathbf{t}_i\}_{i=1}^{m/2},\{\mathbf{z}_i\}_{i=1}^{m/2})\\
    \notag
    &=\sum_{i_{\cdot}=1}^{l} \alpha_{i_{1}} \alpha_{i_{2}}\ldots\alpha_{i_{m/2}} g(\{\mathbf{t}_i\}_{i=1}^{m/2}) \\
    \notag
    &=\sum_{l_{\cdot}=1}^{l}\beta_{l_{1}}\beta_{l_{2}}\ldots\beta_{l_{m/2}} f(\{\mathbf{z}_i\}_{i=1}^{m/2})
\end{align}
where the second equation follows from the definition.
We remark that since the assumption that $\kappa$ is semi-definite, 
\begin{align}
\notag
    \langle f,f \rangle =
    \sum_{i_1,\ldots,i_m} \alpha_{i_1}\ldots \alpha_{i_m} \kappa (\{\mathbf{x}_i\}_{i=1}^{m/2},\{\mathbf{x}_i\}_{i=m/2+1}^{m}) 
    \geq 0.
\end{align}
Similarly to the standard kernel, the biclique kernel has reproducing property.
The reproducing property follows from Eq.~\eqref{eq:innerproduct} if we take $g = \kappa(\{\mathbf{x}_i\}_{i=1}^{m/2}, \cdot)$, and we do the operation defined as Eq.~\eqref{eq:innerproduct} on the function $f$, 
\begin{align}
\notag
    &\langle f, \kappa(\{\mathbf{x}_i\}_{i=1}^{m/2}, \cdot) \rangle \\
\notag
    &= \sum_{i_{\cdot}=1}^{l}\alpha_{i_{1}}\alpha_{i_{2}}\ldots\alpha_{i_{m/2}}\kappa(\{\mathbf{t}_i\}_{i=1}^{m/2},\{\mathbf{x}_i\}_{i=1}^{m/2}) \\
\notag
    &= f(\{\mathbf{x}_i\}_{i=1}^{m/2}).
\end{align}
We call this property as \textit{reproducing property}.

To conclude the proof, it remains to show separability and completeness. 
Since $\kappa$ is also kernel, separability follows for the same reasoning as~\cite{shawe2004kernel}.
For completeness, we consider a fixed input $\{\mathbf{x}_i\}_{i=1}^{m/2}$ and a Cauchy sequence $(f_n)_{n=1}^{\infty}$.
From the Cauchy-Schwarz inequality, we obtain
\begin{align}
\notag
    &\left(f_{n}(\{\mathbf{x}_i\}_{i=1}^{m/2})-f_{m}(\{\mathbf{x}_i\}_{i=1}^{m/2})\right)^{2}\\
\notag
    &=\left\langle f_{n}-f_{m}, \kappa(\{\mathbf{x}_i\}_{i=1}^{m/2}, \cdot)\right\rangle^{2}\\
    \notag
    &\leq\left\|f_{n}-f_{m}\right\|^{2} \kappa(\{\mathbf{x}_i\}_{i=1}^{m/2}, \{\mathbf{x}_i\}_{i=1}^{m/2}).
\end{align}
This means that the Cauchy sequence $(f_n)_{n=1}^{\infty}$ is bounded, and has a limit. 
We define such a limit 
\begin{align}
\notag
    g(\{\mathbf{x}_i\}_{i=1}^{m/2}) = \lim_{n \to \infty} f_{n}(\{\mathbf{x}_i\}_{i=1}^{m/2}),
\end{align}
and include all such limit functions in $\mathcal{F}$. 
Then, we obtain the Hilbert space $\mathcal{F}_{\kappa}$ associated with kernel $\kappa$.

While we so far have the feature space, we need to specify the image of an input $\{\mathbf{x}_i\}_{i=1}^{m/2}$ under the mapping $\psi$.
\begin{align}
\notag
    \psi(\{\mathbf{x}_i\}_{i=1}^{m/2}) = \kappa(\{\mathbf{x}_i\}_{i=1}^{m/2}, \cdot) \in \mathcal{F}_{\kappa}.
\end{align}
Then, inner product between an element of $\mathcal{F}_{\kappa}$ and the image of an input $\{\mathbf{x}_i\}_{i=1}^{m/2}$ is
\begin{align}
\notag
    \langle f, \psi(\{\mathbf{x}_i\}_{i=1}^{m/2}) \rangle = \langle f, \kappa(\{\mathbf{x}_i\}_{i=1}^{m/2}, \cdot) \rangle = f(\{\mathbf{x}_i\}_{i=1}^{m/2}).
\end{align}
This is what we need. 
Furthermore, the inner product is strict since if $\|f\|=0$, then $\forall x$ we have
\begin{align}
\notag
    f = \langle f, \psi(\{\mathbf{x}_i\}_{i=1}^{m/2}) \rangle  \leq \|f\|\|\psi(\{\mathbf{x}_i\}_{i=1}^{m/2})\| = 0.
\end{align}

\section{Proof of Lemma.~\ref{lemma:bicliquekernelmatrixation} and Proposition~\ref{prop:tensorandmatrix}}
Here we provide the proof of Lemma.~\ref{lemma:bicliquekernelmatrixation} and Proposition~\ref{prop:tensorandmatrix}.
Actually, Prop.~\ref{prop:tensorandmatrix} follows directly from the proof of Lemma~\ref{lemma:bicliquekernelmatrixation}.
Here, we firstly discuss the proof for Lemma~\ref{lemma:bicliquekernelmatrixation} then we apply Lemma~\ref{lemma:bicliquekernelmatrixation} to prove Prop.~\ref{prop:tensorandmatrix}.

\subsection{Proof of Lemma~\ref{lemma:bicliquekernelmatrixation}}
We begin the proof by computing a gram matrix $K^{(m)}$ of $\kappa^{(m)}$.
First, to avoid confusion in proof, we omit the variables $\mathbf{t}_{\cdots}$ in the definition.
We rewrite our kernel $\kappa^{(m)}$ for the two sets of $m/2$ variables $\{\mathbf{x}_{i_{\cdot}}\}$ and $\{\mathbf{x}_{i_{m/2+\cdot}}\}$ as 
\begin{align}
     \kappa^{(m)}(\{x_{i_{\mu}}\}_{\mu=1}^{m/2},\{x_{i_{\mu}}\}_{\gamma=m/2+1}^{m}) =\sum_{j_{1}, j_{2}=1}^{m/2} \kappa(\mathbf{x}_{i_{j_{1}}},\mathbf{x}_{i_{i_{m/2+j_{2}}}}),
\end{align}
instead of $\{\mathbf{x}_{i_{\cdot}}\}$ and $\{\mathbf{t}_{l_{\cdot}}\}$ in the original definition Eq.~\eqref{eq:generalizedkernel}. 
This writing change does not change the definition, but just rewrites the variables. 

To save the space, we introduce the abbreviation as 
\begin{align}
\notag
\psi_{i}  &:= \psi(\mathbf{x}_{i})\\
\notag
\psi_{ij} &:= \langle \psi(\mathbf{x}_{i}),\psi(\mathbf{x}_{j}) \rangle.
\end{align}
Let $\kappa^{(m)}$ for two $m/2$ variables $\{\mathbf{x}_{i},\mathbf{x}_{i_{1}},\ldots,\mathbf{x}_{i_{m/2-1}} \}$, $\{\mathbf{x}_{j},\mathbf{x}_{x_{m/2}},\ldots,\mathbf{x}_{m-2} \}$
For $X=\{\mathbf{x}_{1},\ldots,\mathbf{x}_{n}\}$, we can compute the gram matrix as
\begin{align}
\notag
&K^{(m)}_{il} \\
\notag
=& \sum_{i_{\cdot},j_{\cdot}} \kappa^{(m)}(\{\mathbf{x}_i\}, \cup \{\mathbf{x}_{i_{\cdot}}\},\{\mathbf{x}_j\}\cup\{\mathbf{x}_{m/2+j_{\cdot}}\})\\
\notag
&K^{(m)}_{ij} = \sum_{i_{\cdot},j_{\cdot}}\psi_{ij} + \underbrace{\psi_{ij_{1}} + \ldots +\psi_{ij_{m/2-1}}}_{m/2-1 \mathrm{\  terms}} \\
\notag
&+ \underbrace{\psi_{ji_{1}} + \ldots +\psi_{ji_{m/2-1}}}_{m/2-1 \mathrm{\ terms}}\\
\notag
&+\underbrace{\psi_{i_{1}j_{1}} +\psi_{i_{1}j_{2}} + \ldots + \psi_{i_{m/2-1}j_{m/2-1}}}_{(m/2 -1) \times (m/2 -1) \mathrm{\ terms}}\\
\label{eq:grammatrixwrittendown}
=& n^{m-2}\psi_{ij} + n^{m-3}\frac{m-2}{2}\sum_{l=1}^{n}(\psi_{il} + \psi_{jl}) \\
\notag
&+ n^{m-4}\left( \frac{m-2}{2} \right)^2 \sum_{l,k=1}^{n} \psi_{lk} \\
\label{eq:grammatrixkernel}
=& n^{m-2}\langle \psi_{i} + \frac{m-2}{2}\sum_{l=1}^{n}\frac{\psi_{l}}{n},\psi_{j} + \frac{m-2}{2}\sum_{l=1}^{n}\frac{\psi_{l}}{n} \rangle.
\end{align}
Note that $i_{\cdot}$ and $j_{\cdot}$ runs from 1 to $n$.
Then, 
\begin{align}
\sum_{i_{\cdot},j_{\cdot}}\psi_{ij_{r}}&=n^{m-3} \sum_{l} \psi_{il}\\
\sum_{i_{\cdot},j_{\cdot}} \psi_{ji_{r}} &= n^{m-3} \sum_{l}\psi_{jl},
\end{align}
for all $r$$=$$1$$,\ldots,$$m/2$$-$$1$,
\begin{align}
    \sum_{i_{\cdot},j_{\cdot}}\psi_{i_{r}j_{s}}=n^{m-4}\sum_{lk}\psi_{lk}
\end{align} 
for all $r,s$$=$1$,\dots,$$m/2$$-$$1$, and
\begin{align}
\sum_{i_{\cdot},j_{\cdot}}\psi_{ij}=n^{m-2}\psi_{ij}    
\end{align} 
Using this we obtain Eq.~\eqref{eq:grammatrixwrittendown}.
By Eq.~\eqref{eq:grammatrixkernel}, we now prove Lemma.~\ref{lemma:bicliquekernelmatrixation}. 
We remark that Eq.~\eqref{eq:grammatrixwrittendown} is equivalent to Eq.~\eqref{eq:matrixgeneralizedkernel}, since $\psi_{ij}=K_{ij}$ by definition.
Since $\kappa^{'(m)}$ is kernel by Eq.~\eqref{eq:grammatrixkernel}, $K^{(m)}$ is semi-definite, which concludes Cor.~\ref{cor:semidefinite}.

\subsection{Proof of Proposition~\ref{prop:tensorandmatrix}}
It is clear that there exists unique $\kappa^{'(m)}$ from $\kappa^{(m)}$, by seeing the way of composition.
We move on to show that there exists unique $\kappa^{(m)}$ from $\kappa^{'(m)}$ and $m$.
Since $K^{(m)}$ is a kernel by Lemma~\ref{lemma:bicliquekernelmatrixation}, this concludes that $K^{(m)}$ is semi-definite. 
For a semi-definite matrix for $A^{(m)}$, we have a decomposition $\psi'$ by the standard Mercer's theorem.
Then, we have
\begin{align}
\notag
    \psi'_{i} = n^{\frac{m-2}{2}}\psi_{i} + n^{\frac{m-2}{4}}\frac{m-2}{2}\sum_{l=1}^{n}\psi_{l},
\end{align}
where $\psi$ is a feature map for biclique kernel for $\mathcal{A}$.
By the construction, we can rewrite $\psi$ by $\psi'$ by solving the linear equation as
\begin{align}
\notag
\psi' = C \psi,
\end{align}
where $\psi':=(\psi'_{1},\ldots, \psi'_{n})^{\top}$, $\psi:=(\psi_{1},\ldots,\psi_{n})^{\top}$, and
\begin{align}
\notag
C:=n^{\frac{m-4}{2}}\left(
\begin{array}{cccc}
n + \frac{m-2}{2} & \frac{m-2}{2} & \ldots & \frac{m-2}{2} \\
 \frac{m-2}{2} & n + \frac{m-2}{2} & \ldots & \frac{m-2}{2}\\
\vdots &  & \ddots & \vdots\\
 \frac{m-2}{2} & \ldots & \frac{m-2}{2} & n+ \frac{m-2}{2}\\
\end{array}
\right).
\end{align}
By construction, $C$ is a full rank matrix, and therefore we can write as
\begin{align}
\label{eq:decompositionforbicliquekernel}
 \psi =  C^{-1} \psi'.
\end{align}
Therefore, this concludes the proof.

\section{On Equivalence for Three Hypergraph Cuts for Uniform Hypergraph}
\label{sec:onequvalence}
We firstly introduce the contraction way for \textit{any} hypergraph, while the main text only assumes uniform hypergraph.
For the convenience of notation, we begin with definition of degree matrix of edge. 
We define the degree of an edge $e$$\in$$E$ is defined as $|e|$.  
We also define the degree matrix $D_{e}$ whose diagonal elements of edges.
Remark that all the diagonal elements of $D_e$ are $m$ for $m$-uniform hypergraph since we $|e|=m, \forall e \in E$.
Using this notation, we expand the star to any hypergraph as $A_{s}$$:=$$HW_{e}D_{e}^{-1}H^{\top}$.
The clique way, which has two kinds of formulations; edge-normalized way by~\cite{saito2018hypergraph} as 
$A_{nc}$$:=$$H W_{e}(D_{e}$$-$$I)^{-1}H^{\top}$$-$$D_{v}$ 
and edge-unnormalized way by~\cite{rod} as 
$A_{uc}$$:=$$H W_{e}H^{\top}$$-$$D_{uc}$, 
where $D_{uc}$ is a diagonal matrix whose diagonal element is $d_i$$=$$\sum_{j}$$\sum_{e \in E:i,j \in e } w(e)$.

Next, we rewrite uniform hypergraph cut (defined as Eq.~\eqref{eq:defhypergraphcut}) to any hypergraph, as
\begin{align}
\notag
    \mathrm{Cut}(V_{1},V_{2}) &:= \sum_{e\in E} w(e) \frac{|e \cap V_{1}||e \cap V_{2} |}{|e|}\\
    \label{eq:defanyhypergraphcut}
    &=\sum_{e\in E } \sum_{j_1,j_2 \in e; j_1 \in V_{1}, j_2 \in V_{2}} \frac{w(e)}{|e|}.
\end{align}
By computing further, we obtain the same eigenproblem as Prop.~\ref{prop:cutequivalent}.

While Prop.~\ref{prop:cutequivalent} is a cut for $A_{s}$, the cut for the other two $A_{uc}$ and $A_{nc}$ is defined by replacing the denominator of Eq.~\eqref{eq:defanyhypergraphcut} by 1 and $|e|$$-$$1$, respectively~\cite{saito2018hypergraph}.
However, these three produce the same results for uniform hypergraph. 
The reason is seeing Eq.~\eqref{eq:deflaphypereigenproblem}, the cuts are rewritten as eigenproblem of Laplacians $L_{uc}$$:=$$D_{uc}$$-$$A_{uc}$ and $L_{nc}$$:=$$D_{v}$$-$$A_{nc}$, but $L_{uc}$$=$$($$m$$-1)L_{s}$$=$$mL_{nc}$ for uniform hypergraph.
We elaborate more as follows.

For uniform hypergraph, if we consider the hypergraph cut problem for edge-unnormalized clique as in~\cite{saito2018hypergraph} as
\begin{align}
    \label{eq:hypergraphcutuc}
    \mathrm{Cut}(V_{1},V_{2}) &:= \sum_{e\in E} w(e)|e \cap V_{1}||e \cap V_{2} |,
\end{align}
the $k$-way normalized cut corresponds eigenproblem of unnormalized clique problem as 
\begin{align}
\notag
    \min \mathrm{kNCut} = \min Z^{\top}D^{-1/2}_{R} L_{uc} D^{-1/2}_{R} Z \mathrm{s.t.} Z^{\top}Z =1.
\end{align}
If we consider the following hypergraph cut problem for edge-normalized clique as
\begin{align}
    \label{eq:hypergraphcutnc}
    \mathrm{Cut}(V_{1},V_{2}) &:= \sum_{e\in E} w(e)\frac{|e \cap V_{1}||e \cap V_{2} |}{m-1},
\end{align}
the $k$-way normalized cut corresponds eigenproblem of normalized clique problem as
\begin{align}
\notag
    \min \mathrm{kNCut} = Z^{\top} D^{-1/2}_{v} L_{nc} D^{-1/2}_{v} Z \mathrm{s.t.} Z^{\top} Z = 1.
\end{align}
The difference of objective functions Eq.~\eqref{eq:defhypergraphcut}, Eq.~\eqref{eq:hypergraphcutuc} and Eq.~\eqref{eq:hypergraphcutnc} are the constant coefficient.
Therefore, the three formulation would produce the same results.

\section{Proof of Proposition~\ref{prop:cutequivalent}}
By definition of the star adjacency matrix, the matrix can be computed
\begin{align}
\notag
    &(A_s)_{ij} = \sum_{e \in E; ij \in e}\frac{a_{ij}}{m}
\end{align}
Considering the order of edges, this would be
\begin{align}
\notag
    (A_s)_{ij} &= \sum_{e \in E; ij \in e}\frac{a_{ij}}{m}\\
    \label{eq:starwrittendown}
    &=\sum_{e \in E;ij \in e} \left(\frac{m}{2}\right)^2w(e={i,\cdots,j,\cdots})
\end{align}
Eq.~\eqref{eq:starwrittendown} is shown to be equivalent to $A_{g}/m$.
Therefore, the problems Eq.~\eqref{eq:tensorhypergraphcut} and Eq.~\eqref{eq:defadjhypereigenproblem} are equivalent.

\section{Detailed of Weighted Kernel $k$-means}
\label{sec:Detailsof}
Here we show the detailed steps of computation from the naive definition of weighted kernel $k$-means to Eq.~\eqref{eq:kernelobjnorm}.
\begin{align}
    \notag
    &J(\pi_j) \\
    \notag
    =& \sum_{\mathbf{x}_{i} \in \pi_{j}} w(\mathbf{x}_i)(\|\psi(\mathbf{x}_i)\|^2 -2 \langle \psi(\mathbf{x}_{i}), \mathbf{m}_j \rangle + \| \mathbf{m}_j \|^2)\\
    \notag
    =& \sum_{\mathbf{x}_{i} \in \pi_{j}}
    \left(w(\mathbf{x}_i)\kappa(\mathbf{x}_i,\mathbf{x}_i) - 2w(\mathbf{x}_i)\sum_{\bm{\beta} \in \pi_{j}} \frac{w(\bm{\beta})}{s_j}\kappa(\mathbf{x}_j,\bm{\beta}) \right.\\
    \notag
    &+ \left. w(\mathbf{x}_j) \sum_{\bm{\alpha},\bm{\beta} \in \pi_{j}}\frac{w(\bm{\alpha})w(\bm{\beta})}{s_j^2}\kappa(\bm{\alpha},\bm{\beta})\right)\\
    \notag
    =&\sum_{\mathbf{x}_{i} \in \pi_{j}}
    w(\mathbf{x}_i)\kappa(\mathbf{x}_i,\mathbf{x}_i) - 2\sum_{\bm{\alpha},\bm{\beta} \in \pi_{j}} \frac{w(\bm{\alpha})w(\bm{\beta})}{s_j}\kappa(\bm{\alpha},\bm{\beta})
    \\
    \notag
    &+ \frac{\sum_{\mathbf{x}_j \in \pi_j} w(\mathbf{x}_j)}{s_j} \sum_{\bm{\alpha},\bm{\beta} \in \pi_{j}}\frac{w(\bm{\alpha})w(\bm{\beta})}{s_j}\kappa(\bm{\alpha},\bm{\beta})\\
    \notag
    &=\sum_{\mathbf{x}_{i} \in \pi_{j}}
    w(\mathbf{x}_i)\kappa(\mathbf{x}_i,\mathbf{x}_i) -
    \sum_{\bm{\alpha},\bm{\beta} \in \pi_j} \frac{w(\bm{\alpha})w(\bm{\beta})}{s_j}\kappa(\bm{\alpha},\bm{\beta}).
\end{align}
Summing up this to all the cluster, we obtain
\begin{align}
\notag
    J(\{\pi\}_{i=1}^{k})=&\sum_{j=1}^{k}\sum_{\mathbf{x}_{i} \in \pi_{j}}
    w(\mathbf{x}_i)\kappa(\mathbf{x}_i,\mathbf{x}_i) \\
\notag
    &-\sum_{\bm{\alpha},\bm{\beta} \in \pi_j}  \left(\frac{w^{1/2}(\bm{\alpha})}{s^{1/2}_j}\right)^2\kappa(\bm{\alpha},\bm{\beta})\left(\frac{w^{1/2}(\bm{\beta})}{s^{1/2}_j}\right)^2\\
    \notag
    =& \mathrm{trace}W^{1/2}KW^{1/2} - \mathrm{trace}YW^{1/2}KW^{1/2}Y.
\end{align}
This concludes the steps.

\section{Details of Heat Kernel for Biclique Kernel}
\label{sec:detailsofheatkernels}
In this section, we discuss the omitted details of Heat Kernel Section of the main text.
\subsection{On Discrete Laplacian Eq.~\eqref{eq:laplaciantensor}}
We start to define a matrix $L^{(m)}$ to satisfy
\begin{align}
    L^{(m)}f'(\mathbf{x}_{1}) = \sum_{\{\mathbf{x}_{i_{\cdot}}\}_{\mu=2}^{m/2}}L^{(m)}_{t,n} f(\{\mathbf{x}_{i_{\cdot}}\}).
\end{align}
Then, we can rewrite the minimization problem as
\begin{align}
\notag
    S_{2}^{(m)}(f)&:=\sum_{\{\mathbf{x}_{i_{\cdot}}\},\{\mathbf{y}_{i_{\cdot}}\}}L^{(m)}_{t,n} f(\{\mathbf{x}_{i_{\cdot}}\})f(\{\mathbf{y}_{i_{\cdot}}\}) \\
    \label{eq:energydiscrete}
    &= \sum_{\mathbf{x},\mathbf{y}} L^{(m)}f'(\mathbf{x})f'(\mathbf{y})
\end{align}
Let $A$ be a contracted matrix from a gram tensor $\mathcal{H}$ for the biclique kernel $H^{(m)}_{t}$, and $D$ be a degree matrix of $A$.
By construction of $L^{m}$, we can rewrite
\begin{align}
\notag
    L^{(m)} = \frac{D}{m/2} - A. 
\end{align}
Now we consider to introduce normalizing constrains. 
This is justified since the continuous counterpart of energy minimization problem Eq.~\eqref{eq:stokes} we can introduce such a constraints by properly choosing the measure $\mu$ for inner product.
Now, we consider Eq.~\eqref{eq:energydiscrete}.
Introducing the normalizing constraints, we write as
\begin{align}
\notag
    \min S_{2}^{(m)}(f) &= \min f^{\top}L^{(m)}f\\
\notag
    &=\min f^{\top} D^{-1/2}(\frac{D}{m/2} - A) D^{-1/2}f\\
\notag
    &=\max f^{\top} D^{-1/2}AD^{-1/2}f, \mathrm{\ s.t.\ } f^{\top}f = 1.
\end{align}
This corresponds to the hypergraph cut problem for the hypergraph formed by $H^{(m)}_{t}$.

\subsection{Proof of the statement that $H_t^{(m)}$ Is a Heat Kernel}
\label{sec:proofHtmisaheatkernel}
Since $H$ is a heat kernel $H:\mathcal{M}\times\mathcal{M}\times (0,\infty) \rightarrow \mathbb{R}$, we can decompose into
\begin{align}
    \label{eq:heatkerneldecomposition}
    H_{t}(x,y) = \sum_{i}\exp(-\lambda_{i}t) \phi_{i}(x)\phi_{i}(y)
\end{align}
Using the fact that $H_t$ is a heat kernel, we can prove that the kernel $H_t^{(m)}$ is a heat kernel $\mathcal{M}^{m/2}\times\mathcal{M}^{m/2}\times (0,\infty) \rightarrow \mathbb{R}$, since we can rewrite $H_t^{(m)}$ as
\begin{align}
    \label{eq:generalizedheatkernldecomposition}
    &H^{(m)}_{t} (\{x_{\cdot}\},\{y_{\cdot}\}) = \\
    \notag
    &\sum_{i} \exp(-\lambda_{i}t) (\phi_{i}(x_{1}) + \ldots \phi_{i}(x_{m/2}))(\phi_{i}(y_{1}) + \ldots \phi_{i}(y_{m/2})).
\end{align}
Similarly to Eq.~\eqref{eq:generalizedheatkernldecomposition}, this heat kernel $H^{'(m)}$ is also a heat kernel since we can rewrite this as
\begin{align}
\notag
    &H^{'(m)}(x,y) \\
    \notag
    =& \sum_{i} \exp(- \lambda_{i}t) (\phi(x) + (m/2-1)\int_{\mathcal{M}'}\phi(x)dx)\\
    \notag
    &\times (\phi(y) + (m/2-1)\int_{\mathcal{M}'}\phi(y)dy).
\end{align}

\subsection{Detailed Steps of Approximation}

This section discusses the details of approximation of asymptotic heat equation.
In this subsection about approximation, we approximate $\mathcal{M}'\approx\mathbb{R}^{n}$, and  $d\mu(y_{\cdot}) \approx dy$, similar to~\cite{belkin2003laplacian}.

We start with expanding the heat equation as done in the main text.
\begin{align}
\notag
    &\lim_{t \to 0} \Delta U(t,\{x_i\}_{i=1}^{m/2}) = \Delta f(\{x_i\}_{i=1}^{m/2})\\ 
    \notag
    &= - \lim_{t \to 0} \frac{\partial}{\partial t} \int_{\mathcal{M}'} d\mu(y_{*})H_{t}(\{x_i\}_{i=1}^{m/2},\{y_{\cdot}\}) U(0,\{y_{\cdot}\})
\end{align}
Therefore, for small $t$, we obtain 
\begin{align}
\notag
     &- \lim_{t \to 0} \frac{\partial}{\partial t} \int_{\mathcal{M}'}d\mu(y_{*}) H_{t}(\{x_i\}_{i=1}^{m/2},\{y_{\cdot}\}) U(0,\{y_{\cdot}\}) \\
     \notag
     & \approx -\frac{1}{t}\left( \int_{\mathcal{M}'} d\mu(y_{*}) H_{t}(\{x_i\}_{i=1}^{m/2},\{y_{\cdot}\}) f(\{y_{\cdot}\}) \right.
     \\
     \label{eq:primitiveapproximation}
     &- \left. \int_{\mathcal{M}'} d\mu(y_{*}) H_{0}(\{x_i\}_{i=1}^{m/2},\{y_{\cdot}\}) f(\{y_{\cdot}\}) \right),
\end{align}
following from the definition of partial differentiation.

In the following, we consider $H_{t} = H^{(m)}_{t}$ since we are interested in this case.
Note that the heat kernel $H^{(m)}_{t}$ is a biclique kernel whose base kernel is Gaussian kernel $G_{t}$. 
To further consider Eq.~\eqref{eq:primitiveapproximation}, we next examine a solution to Eq.~\eqref{eq:heatequation} in the asymptotic case $t \to 0$. 
This allows us to have $\Delta f$, which we want to analyze in Eq.~\eqref{eq:stokes}. 
We provide an analysis on the solution to Eq.~\eqref{eq:heatequation} when $t$$\to$$0$.
\begin{proposition}
    \label{prop:H0}
    In the Euclidean space, i.e., $\mathcal{M}'=\mathbb{R}^{n}$ and $d\mu(y_{i})=dy_{i}$, a given a function $f:\mathcal{M}'\to \mathbb{R}$ and the constraints in Eq.~\eqref{eq:stokes}, then 
    \begin{align}
    \notag
    &\lim_{t \to 0}\int_{\mathcal{M}'} d\mu(y_{*}) H_{t}^{(m)}(\{x_{\cdot}\},\{y_{\cdot}\})f(\{y_{\cdot}\})\\
    \label{eq:H0}
    &=
    \lim_{t \to 0}\int_{\mathcal{M}'} d\mu(y_{*}) \frac{H_{t}^{(m)}(\{x_{\cdot}\},\{y_{\cdot}\})}{m/2}f(\{x_{\cdot}\})
    \end{align}
\end{proposition}
This proposition is a generalized version of Gaussian kernel features in the following sense. For a single variable Gaussian Kernel, we have
\begin{align}
\label{eq:gaussianonef}
    &\lim_{t \to 0}\int_{\mathcal{M'}} G_{t}(x,y) f(y) d\mu(y) = f(x),\\
\label{eq:gaussianone1}
    &\lim_{t \to 0}\int_{\mathcal{M'}} G_{t}(x,y) d\mu(y) = 1.
\end{align}
Combining these two, we obtain
\begin{align}
\notag
    \lim_{t \to 0} \int_{M} G_{t}(x,y) f(y) d\mu(y) &= f(x) \\
\label{eq:singleH0}
    &= \lim_{t \to 0}\int_{M} G_{t}(x,y) f(x) d\mu(y).
\end{align}

Note the difference between the variables of $y$ of $f$ on the left-hand side and $x$ of $f$ on the right-hand side.
Prop.~\ref{prop:H0} is a generalized version of this relationship Eq.~\eqref{eq:singleH0}.

Using this relation Eq.~\eqref{eq:singleH0}, for small $t$, we can further rewrite as
\begin{align}
\notag
    \int_{M} G_{t\to 0}(x,y) f(y) d\mu(y)  &= \lim_{t \to 0} \int_{M} G_{t}(x,y) f(y) d\mu(y) \\
\notag
    & = f(x)\\
\notag
    &= \lim_{t \to 0}\int_{M} G_{t}(x,y) f(x) d\mu(y)\\
\notag
    &= \int_{M} G_{t\to 0}(x,y) f(x) d\mu(y).\\
\notag
    &\approx \int_{M} G_{t}(x,y) f(x) d\mu(y).
\end{align}
It is too rough to approximate the left-hand side in the form of the right-hand side, i.e.,
\begin{align}
\notag
    \lim_{t \to 0} \int_{M} G_{t}(x,y) f(y) d\mu(y) \not \approx \int_{M} G_{t}(x,y) f(y) d\mu(y)
\end{align}
The reason is that since we have the variable $y$, which we take integral in both the Gaussian and the function, we cannot see the approximated shape if we increase the value of $t$, even if $t$ is very small. 
The right-hand side overcomes this problem.
Using this relation, a single variable version of Eq.~\eqref{eq:primitiveapproximation} is further approximated.

Now, using the same strategy for Eq.~\eqref{eq:H0} in Prop.~\ref{prop:H0}, we have an approximation for small $t$ as
\begin{align}
    \notag
    &\int_{\mathcal{M}'} d\mu(y_{*}) H_{t\to 0}^{(m)}(\{x_{\cdot}\},\{y_{\cdot}\})f(\{y_{\cdot}\})\\
    \notag
    &=\lim_{t \to 0}\int_{\mathcal{M}'} d\mu(y_{*}) H_{t}^{(m)}(\{x_{\cdot}\},\{y_{\cdot}\})f(\{y_{\cdot}\})\\
    \notag
    &=
    \lim_{t \to 0}\int_{\mathcal{M}'} d\mu(y_{*}) \frac{H_{t}^{(m)}(\{x_{\cdot}\},\{y_{\cdot}\})}{m/2}f(\{x_{\cdot}\})\\
    \notag
    &=\int_{\mathcal{M}'} d\mu(y_{*}) \frac{H_{t\to 0}^{(m)}(\{x_{\cdot}\},\{y_{\cdot}\})}{m/2}f(\{x_{\cdot}\}) \\
    \label{eq:approximatedrelationformulti}
    &\approx \int_{\mathcal{M}'} d\mu(y_{*}) \frac{H_{t}^{(m)}(\{x_{\cdot}\},\{y_{\cdot}\})}{m/2}f(\{x_{\cdot}\})
\end{align}
This relation further rewrites Eq.~\eqref{eq:primitiveapproximation}, as    
\begin{align}
\notag
     &- \lim_{t \to 0} \frac{\partial}{\partial t} \int_{\mathcal{M}'} d\mu(y_{*})H_{t}(\{x_i\}_{i=1}^{m/2},\{y_{\cdot}\}) U(0,\{y_{\cdot}\}) \\
     \notag
     & \approx -\frac{1}{t}\left( \int_{\mathcal{M}'} d\mu(y_{*}) H_{t}(\{x_i\}_{i=1}^{m/2},\{y_{\cdot}\}) f(\{y_{\cdot}\}) \right.
     \\
     \notag
     &- \left. \int_{\mathcal{M}'} d\mu(y_{*}) H_{0}(\{x_i\}_{i=1}^{m/2},\{y_{\cdot}\}) f(\{y_{\cdot}\}) \right)
    \\
    \notag
    &\approx -\frac{1}{t}\left( \int_{\mathcal{M}'} d\mu(y_{*}) H_{t}(\{x_{\cdot}\},\{y_{\cdot}\}) f(\{y_{\cdot}\}) \right.
     \\
     \label{eq:primitiveapproximationfurther}
     &- \left. \int_{\mathcal{M}'} d\mu(y_{*}) \frac{H_{t}^{(m)}(\{x_{\cdot}\},\{y_{\cdot}\})}{m/2}f(\{x_{\cdot}\}) 
     \right).
\end{align}
To obtain the third approximation, we apply Eq.~\eqref{eq:approximatedrelationformulti} to the second term of the second equation. We also use the fact that $\{x_{\cdot}\}$ is an abbreviated form of $m/2$ continuous variables $\{x_i\}_{i=1}^{m/2}$.

Consider discrete data points in $\mathcal{M}'$ instead of the continuous variables, 
Eq.~\eqref{eq:primitiveapproximation}, which is approximated as Eq.~\eqref{eq:primitiveapproximationfurther}, is further approximated by
\begin{align}
\notag
      &\frac{1}{t}\left( -\sum_{\{\mathbf{y}_{i_{\cdot}}\}} H_{t}^{(m)}(\{ \mathbf{x}_{i_{\cdot}} \},\{\mathbf{y}_{i_{\cdot}}\})f(\{ \mathbf{y}_{i_{\cdot}} \})\right.
       \\
\notag
      &
      \left.+\sum_{\{ \mathbf{y}_{i_{\cdot}} \}} \frac{H_{t}^{(m)}(\{ \mathbf{x}_{i_{\cdot}} \},\{\mathbf{y}_{i_{\cdot}}\})}{m/2}f(\{\mathbf{x}_{i_{\cdot}}\})
      \right) = \frac{1}{t} L^{(m)}_{t,n} f(\{\mathbf{x}_{i_{\cdot}}\})
\end{align}
Thus, the Laplacian Eq.~\eqref{eq:laplaciantensor} can be seen as a discrete approximation of the continuous Laplacian for $m/2$ variables heat equation Eq.~\eqref{eq:heatequation}. 
We also see that Prop.~\ref{prop:H0} introduces the coefficient $m/2$ in Eq.~\eqref{eq:laplaciantensor}.

Finally, we remark that all the approximation here is justified by Thm.~\ref{thm:lapalcianconverge}.

\subsection{Proof of Proposition~\ref{prop:H0}}
\label{sec:proofofpropH0}
This section provides a proof of Prop.~\ref{prop:H0}.

Before we proceed, we review the one variable case.
Using single variable Gaussian kernel characteristics Eq.~\eqref{eq:gaussianonef} and Eq.~\eqref{eq:gaussianone1}, we compute for multivariate version of Eq.~\eqref{eq:gaussianonef} as
\begin{align}
\notag
    &\lim_{t \to 0} \int_{\mathcal{M'}^{m/2}} H^{(m)}_{t}(\{x_i\}_{i=1}^{m/2},\{y_i\}_{i=1}^{m/2}) f(y) d\mu(y_{*}) \\
    \notag
    &= \lim_{t \to 0} \int_{\mathcal{M'}^{m/2}} \sum_{i=1}^{m/2} \sum_{j=1}^{m/2} G_{t}(x_{i}, y_{i}) \sum_{i=1}^{m/2} f'(y_{i})  d\mu(y_{*})\\
    \label{eq:gaussianmultif}
    &= \mathrm{vol}(\mathcal{M})^{m/2-1} \left( \frac{m}{2}\right)^2\sum_{i=1}^{m/2} f'(y_{i}),
\end{align}
and for multivariate version of Eq.~\eqref{eq:gaussianone1} as
\begin{align}
\notag
    &\lim_{t \to 0} \int_{\mathcal{M'}^{m/2}} H^{(m)}_{t}(\{x_i\}_{i=1}^{m/2},\{y_i\}_{i=1}^{m/2}) d\mu(y_{*}) \\
    \notag
    &= \lim_{t \to 0} \int_{\mathcal{M'}^{m/2}} \sum_{i=1}^{m/2} \sum_{j=1}^{m/2} G_{t}(x_{i}, y_{i}) d\mu(y_{*})\\
    \notag
    &= \mathrm{vol}(\mathcal{M})^{m/2-1} \left( \frac{m}{2}\right)^2.
\end{align}
We now prepare to expand the left hand side of Eq.~\eqref{eq:H0}. 
The left hand side of Eq.~\eqref{eq:H0} is expanded as
\begin{align}
\notag
&\lim_{t \to 0}
\int_{\mathcal{M}'} d\mu(y_{*}) H_{t}(\{x_i\}_{i=1}^{m/2},\{y_{\cdot}\})f(\{y_{\cdot}\})\\
\label{eq:heatkernelsolutionbreakdown}
    &= \lim_{t \to 0} \sum_{i=1}^{m/2}\left( \int_{\mathcal{M'}^{m/2}} H^{(m)}_{t}(\{x_i\}_{i=1}^{m/2},\{y_i\}_{i=1}^{m/2})f'(y_{i}) d\mu(y_{*})\right)
\end{align}

We further compute the inside of the parenthesis in Eq.~\eqref{eq:heatkernelsolutionbreakdown} as
\begin{align}
\notag
&\lim_{t \to 0} \int_{\mathcal{M'}^{m/2}} H^{(m)}_{t}(\{x_i\}_{i=1}^{m/2},\{y_i\}_{i=1}^{m/2})f'(y_{i}) d\mu(y_{*}) \\
\notag
=& 
\lim_{t \to 0}\sum_{j=1}^{m/2} \left(\int_{\mathcal{M'}^{m/2}} G_{t}(x_{j},y_{i})f'(y_{i}) d\mu(y_{*}) \right. \\
\notag
&+ \left. \sum_{l\neq i}\int_{\mathcal{M'}^{m/2}} G_{t}(x_{j},y_{l}) f'(y_{i}) d\mu(y_{*}) \right)\\
\notag
=& \sum_{j=1}^{m/2} \left( \mathrm{vol}(\mathcal{M})^{m/2-1} f'(x_{j}) \right.\\
\notag
&+\left.\frac{m/2-1}{2} \mathrm{vol}(\mathcal{M})^{m/2-1} \int_{\mathcal{M}}d\mu(y_{i})f'(y_{i}) \right)\\
\notag
=& \mathrm{vol}(\mathcal{M})^{m/2-1} \sum_{j=1}^{m/2} f'(x_{j}) \\
\notag
&+ \frac{m}{2}\frac{m/2-1}{2} \mathrm{vol}(\mathcal{M})^{m/2-1} \int_{\mathcal{M}}\mu(y_{i})f'(y_{i})
\end{align}
Putting this into Eq~\eqref{eq:heatkernelsolutionbreakdown}, we obtain
\begin{align}
\notag
    &\lim_{t \to 0} \sum_{i=1}^{m/2}\left( \int_{\mathcal{M'}^{m/2}} H^{(m)}_{t}(\{x_i\}_{i=1}^{m/2},\{y_i\}_{i=1}^{m/2})f'(y_{i}) d\mu(y_{*})\right)\\
    \notag
    =&\sum_{i=1}^{m/2}\left(\mathrm{vol}(\mathcal{M})^{m/2-1} \sum_{j=1}^{m/2} f'(x_{j}) \right. \\
    \notag
    &+ \left. \frac{m}{2}\frac{m/2-1}{2} \mathrm{vol}(\mathcal{M})^{m/2-1} \int_{\mathcal{M}}d\mu(y_{i})f'(y_{i})\right)\\
    \notag
    =&\frac{m}{2}\mathrm{vol}(\mathcal{M})^{m/2-1} \sum_{i=1}^{m/2}f'(x_{i}) \\
    \notag
    &+ \frac{m}{2}\frac{m/2-1}{2} \mathrm{vol}(\mathcal{M})^{m/2-1} \sum_{i=1}^{m/2}\int_{\mathcal{M}}d\mu(y_{i})f'(y_{i})\\
    \label{eq:proph0conclusion}
    =& \lim_{t \to 0}\int_{\mathcal{M'}}d\mu(y_{*}) \frac{H^{(m)}_{t}(\{x_i\}_{i=1}^{m/2},\{y_i\}_{i=1}^{m/2})}{m/2} f(\{x_i\}_{i=1}^{m/2}).
\end{align}
The last equality follows from that due to Eq.~\eqref{eq:gaussianmultif} we can proceed the first term as this, and that the second term vanishes due to the constraint $\langle f, c\mathbf{1} \rangle=0$, i.e.,
\begin{align}
\notag
    0 &= \langle f, c\mathbf{1} \rangle\\
\notag
    &= c\int_{\mathcal{M}^{m/2}}d\mu(y_{*})f(\{y_{\cdot}\})\\
\notag
    &= c\int_{\mathcal{M}^{m/2}}\prod_{i=1}^{m/2} d\mu(y_{i})\left(\sum_{i=1}^{m/2}f'(y_{i})\right)\\
\notag
    &= c\sum_{i=1}^{m/2}\mathrm{vol}(\mathcal{M})^{m/2-1} \int_{\mathcal{M}}d\mu(y_{i})f'(y_{i}).
\end{align}
Eq.~\eqref{eq:proph0conclusion} concludes the proof.

\section{Proof of Theorem~\ref{thm:lapalcianconverge}}
\label{sec:prooflaplacianconv}

The strategy to prove Thm.~\ref{thm:lapalcianconverge} is using Hoeffding's inequality.
We start by reviewing Hoeffding's inequality.
\begin{lemma}[Hoeffding]
\label{lemma:hoeffding}
Let $X_1,\ldots,X_n$ be independent identically distributed random variables, such that $|X_{i}| \leq K$. Then
\begin{align}
    P 
    \left( 
    \left|\sum_{i}\frac{X_{i}}{n} - E(X_i)
    \right| 
    >  \epsilon
    \right)
    <
    2 \exp \left( - \frac{\epsilon^2 n}{2K^{2}} \right)
\end{align}
\end{lemma}

To prove the theorem for $L^{(m)}_{tn}$, we evaluate the equation in Eq.~\eqref{eq:laplaciantensor}.
We define the operator $L_{t}^{(m)}:L^{2}(\mathcal{M}) \to L^{2}(\mathcal{M}$ as
\begin{align}
\notag
L^{(m)}_{t} f(\{\mathbf{x}_{i_{\cdot}}\}) 
&:= \int_{\mathcal{M}'} d \mu(y_{*}) \frac{H_{t}^{(m)}(\{\mathbf{x}_{i_{\cdot}}\},\{\mathbf{y}_{i_{\cdot}}\})}{m/2}f( \{\mathbf{x}_{i_{\cdot}}\} \\
\label{eq:tapproximationofoperator}
&- \int_{\mathcal{M}'} d \mu(y_{*}) H_{t}^{(m)} (\{\mathbf{x}_{i_{\cdot}}\},\{\mathbf{y}_{i_{\cdot}}\})f(\{\mathbf{y}_{i_{\cdot}}\})
\end{align}
We remark that $L^{(m)}_{t}$ is the empirical average of $n$ independent random variables with the expectation
\begin{align}
\notag
    E(L_{nt}^{(m)} f(\{\mathbf{x}_{i_{\cdot}}\})) = L_t^{(m)} f(\{\mathbf{x}_{i_{\cdot}}\}).
\end{align}
Applying Hoeffding inequality (Lemma~\ref{lemma:hoeffding}), we obtain
\begin{align}
\notag
    P \left( \left| \frac{1}{t} \frac{L_{n}^tf(\{\mathbf{x}_{i_{\cdot}}\})}{n} - L_{t}^{(m)} f(\{\mathbf{x}_{i_{\cdot}}\}) \right| > \epsilon  \right) \leq \exp \left(  - \frac{\epsilon^2n t^2}{2}\right).
\end{align}
If we choose $t$ as a function of $n$, letting $t = t_{n} = n^{-1/(2+\alpha)}$ where $\alpha >0$ to the equation Thm.~\ref{thm:lapalcianconverge}, we can obtain for any $\epsilon >0$,
\begin{align}
\notag
    &\lim_{n \to \infty} P \left(\left|\frac{1}{t_{n}} \frac{L_{nt}^{(m)}}{n} f(\{\mathbf{x}_{i_{\cdot}}\}) - L^{(m)}_{t} f(\{\mathbf{x}_{i_{\cdot}}\}) \right| > \epsilon  \right)\\
    \notag
    &\leq \lim_{n \to \infty} \exp \left(  - \frac{\epsilon^2 n t_{n}^{2} }{2}\right)\\
    \notag
    &\leq \lim_{n \to \infty}  \exp \left(  - \frac{\epsilon^2 n (n^{-\frac{1}{2+\alpha}})^2 }{2}\right)\\
    \notag
    & =0.
    \end{align}
With the discussion in the proof of Prop.~\ref{prop:H0}, 
we can see that
\begin{align}
\notag
    \lim_{t} L^{(m)}_{t} f(\{\mathbf{x}_{i_{\cdot}}\}) = \Delta f(\{\mathbf{x}_{i_{\cdot}}\}).
\end{align}
If $n \to \infty$ then, $t \to 0$.
The above discussion in all leads the conclusion,
\begin{align}
\notag
    \lim_{n \to \infty} \frac{C}{n t_{n}} L_{n}^{t_{n}} f(\{x_{\cdot}\}) = \Delta f(\{x_{\cdot}\}).
\end{align}

\section{Connection to the Symmetric Modeling and Inhomogeneous Cut}

This section discusses connections from our theoretical properties to the symmetric modeling and inhomogeneous cut. 
If we further assume something on biclique kernel, we may lose a kernel property, i.e., the multi-way relationship may lose a tensor semi-definite property.
Since our discussion fully exploit this property, not every theoretical property holds if we extend from half-symmetric biclique kernel to something beyond that does have a biclique kernel property. However, it is worth looking at what remains. 

Let us look at the connection with weighted kernel $k$-means.
If we only look at the connection, we use the relationship between the tensor and its contracted matrix. 
At the same time, at the both ends, we use kernel property; at the tensor end we use biclique kernel, and at the contracted matrix end we also use a kernel property for the gram matrix of the biclique kernel. 
The discussion on the multivariate heat equation and heat kernel hugely depends upon the biclique kernel property. 
This means that once we lose the the biclique kernel property, we immediately lose the assumption of the discussion of multivariate heat equation.
Thus, in the following, we focus on the connection between weighted kernel $k$-means and these additional modelings.

\subsection{Symmetric Modeling}

This section explores symmetric modeling of the weighted kernel $k$-means.

\subsubsection{Tensor and Weighted Kernel $k$-means Objective Function}
This section discusses the connection between tensor and the weighted kernel $k$-means objective function.

In the main text, we generalized weighted kernel $k$-means to our biclique kernel, see Eq.~\eqref{eq:defgeneralizedkernelkmeans}.
Here, we further consider to generalize to a general symmetric tensor.
The strategy is as follows.
Recall that Eq.~\eqref{eq:kernelobj} is a  gram matrix form of weighted kernel $k$-means objective function Eq.~\eqref{eq:kernelkmeans}.
Recall also that the biclique weighted kernel $k$-means  Eq.~\eqref{eq:defgeneralizedkernelkmeans} is a generalization Eq.~\eqref{eq:kernelkmeans}.
Observing these two objective function, Eq.~\eqref{eq:kernelobj} and Eq.~\eqref{eq:defgeneralizedkernelkmeans}, we replace these kernel related objects, the gram matrix and the gram tensor of biclique kernel, to the general tensor.
We now formalize this discussion.
We consider a symmetric tensor $\mathcal{A}$.
The tensor we are considering is beyond biclique kernel.
Thus, the naming weighted ``kernel'' $k$-means is not appropriate; instead, we use \textit{weighted tensor $k$-means}.
In the following, we replace our biclique kernel in Eq.~\eqref{eq:defgeneralizedkernelkmeans} to the symmetric tensor, as
\begin{align}
\notag
    &J'(\{\pi_j\}_{j=1}^{k}) :=
    \sum_{i \in \pi_{j},j}
    \sum_{
    \{i_{\cdot}\}
    \subset \pi_{j}
    }
    w_{i}\mathcal{A}_{i,i,i_{\cdot}} \\ 
    \label{eq:deftensorkmeans}
    &
    -\sum_{i,l \in \pi_j,j}
    \sum_{
    \{i_{\cdot}\} \subset \pi_{j}
    } \frac{w_{i}w_{l}\mathcal{A}_{i,l,i_{\cdot}}}{s_{j}},\\
\label{eq:weightedtensorkmeanscontraction}
&= \mathrm{trace}WA_{g}W - \mathrm{trace}YW^{1/2}A_{g}W^{1/2}Y,
\end{align}
where $Y$ is an indicator matrix defined as Eq.~\eqref{eq:multipleindicator} and $A_{g}$ is Eq.~\eqref{eq:contractiongd}. 
Seeing Eq.~\eqref{eq:weightedtensorkmeanscontraction}, the objective function of the weighted tensor $k$-means is that solving the spectral clustering for $A_{g}$.
Thus, the weighted tensor $k$-means objective Eq.~\eqref{eq:deftensorkmeans} is equivalent to the spectral clustering proposed by~\cite{ghoshdastidar2015provable}.

\subsubsection{Symmetric Modeling for Even $m$}

If we want a symmetric modeling, instead of half-symmetric modeling in the main text, a simple modification would lead to this.
Remark that if a tensor is symmetric it is half-symmetric. This means that half-symmetricity is a larger concept than symmetricity. 
In this response we write the data point in $\mathbf{X}$ as $x_{i}$instead of $\mathbf{x}_{i}$.
For even $m$ variables $(\mathbf{x}_{1}, \ldots, \mathbf{x}_{m}) \in \mathbf{X}$, we can model symmetric multi-way modeling $\kappa^{(m)}_{s}(\mathbf{x}_{1},\ldots, \mathbf{x}_{m})$ using the current (half-symmetric) biclique kernel $\kappa^{(m)}$ as
\begin{align}
\label{eq:defsymmetricbefore}
\kappa_{s}^{(m)} (\mathbf{x}_{1},\ldots, \mathbf{x}_{m} )&:= \frac{1}{m^m}\sum_{i_1,\ldots,i_m=1}^{m,\ldots,m} \kappa^{(m)}(\mathbf{x}_{i_1},\ldots, \mathbf{x}_{i_{m}})\\
\label{eq:defsymmetric}
&= \frac{1}{4} \sum_{i=1}^{m}\sum_{j=1}^{m} \kappa (\mathbf{x}_i, \mathbf{x}_j)
\end{align}
where $\kappa$ is a base kernel for $\kappa^{(m)}$.
We obtain Eq.~\eqref{eq:defsymmetric} as follows. 
For each biclique kernel, we have a summation of $(m/2)^{2}$ base kernels. 
If we sum in a way like Eq.~\eqref{eq:defsymmetric}, we sum $m^m$ biclique kernels. 
Thus, before we simplify Eq.~\eqref{eq:defsymmetricbefore}, we have $(m/2)^{2} m^m$ terms of base kernels.
The sum of the base kernel $\sum_{i=1}^{m}\sum_{j=1}^{m} \kappa (\mathbf{x}_i, \mathbf{x}_j)$ has $m^2$ terms.
Furthermore, from symmetricity, Eq.~\eqref{eq:defsymmetricbefore} contains the same number of base kernels of each pair.
Hence, it is easy to say that Eq.~\eqref{eq:defsymmetricbefore} can be simplified as Eq.~\eqref{eq:defsymmetric}.

Consider to contract a tensor formed by $\kappa_{s}^{(m)}$ to a matrix $K_s^{(m)}$ by Eq. (3). Then, if $m \geq 4$, $K_s^{(m)}$ is a gram matrix of the kernel $\kappa'_{s}$ defined as
\begin{align}
\label{eq:symmetricgrammatrix}
 \kappa'_{s} (\mathbf{x}_{i}, \mathbf{x}_{j}) :=&\frac{n^{m-2}}{4} \bigl\langle \psi(\mathbf{x}_{i}) + (m-1) \sum_{l=1}^n\frac{\psi(\mathbf{x}_l)}{n} , \\ 
&\psi(\mathbf{x}_{j}) + (m-1) \sum_{l=1}^n\frac{\psi(\mathbf{x}_l)}{n} \bigr\rangle_{\kappa},
\end{align}

This means that $K_{s}^{(m)}$ is a gram matrix for a kernel given $\sum_{l}\psi(\mathbf{x}_l)/n$. For this kernel, we have following indications; i) Since the contraction forms a kernel, the same theoretical discussions like the weighted kernel $k$-means.
ii) Since the kernel is a scale and constant shift variant from the gram matrix for the biclique kernel (See Eq.~\eqref{eq:generalizedkernel}), the experimental result will be very similar to the results of the biclique kernel because scale and constant shift operation will not drastically change the topology.

However, we note that this $\kappa_{s}^{(m)}$ is not a biclique kernel. Therefore, an argument like Thm.~\ref{thm:kernelsemidefinite} does not hold. The reason why 
$\kappa_{s}^{(m)}$ is not a biclique kernel is the same as the reason why the following example is not a kernel.
Consider the $m=2$ case. Then the symmetric similarity function is
 $\kappa_{s}^{(2)}(x_1,x_2) = 2\kappa(x_1,x_2) +  \kappa(x_1, x_1) + \kappa (x_2, x_2)$.
We see that this is not a kernel. 

\subsubsection{Symmetric Modeling for Odd $m$}

We currently do not have semi-definiteness or related concepts for odd order tensors. 
Thus, at this point, we do not have any tools to connect the odd order tensor to the kernel-like discussion like one in the main text. 
We cannot provide Eq.~\eqref{eq:kmeansobjectiveforbicliquekernel}-style objective function for the weighted kernel $k$-means or cannot generalize Thm.~\ref{thm:lapalcianconverge}.

Rather, conversely, we can put any relation into odd order tensor. The example for modeling for odd order uniform hypergraph can be a generalization of Eq.\eqref{eq:defsymmetric}, as
\begin{align}
\label{eq:defsymmetricodd}
    \kappa_{s}^{(m)} (\mathbf{x}_{1},\ldots, \mathbf{x}_{m} )&:= \frac{1}{4}\sum_{i=1}^{m}\sum_{j=1}^{m} \kappa (\mathbf{x}_i,\mathbf{x}_j),
\end{align}
which can be consistent with the even symmetric modeling Eq.~\eqref{eq:defsymmetric}.

However, needless to say, we lose the sense of the kernel.
Further from the even case, even the contraction matrix is not semi-definite.
Thus, we need to compute the contraction naively; instead of having a faster way to compute the contraction, such as Eq.~\eqref{eq:matrixgeneralizedkernel};

\subsubsection{Takeaway from Symmetric Modeling}

So far, we see the symmetric modeling of $m$-uniform hypergraph.
If we take a path to the symmetric modeling, we immediately lose theoretical support from heat equation discussion Thm.~\ref{thm:lapalcianconverge}, which is a generalized version of~\cite{belkin2003laplacian}.
If we focus on the $m$ even case, the contracted matrix still have semi-definiteness. 
Thus, we can see some properties of kernel, in the sense of the weighted kernel $k$-means; we obtain a similar representation of weighted kernel $k$-means objective like Eq.~\eqref{eq:kmeansobjectiveforbicliquekernel}.
Moreover, using Eq.~\eqref{eq:symmetricgrammatrix}, we obtain the contracted matrix faster.
Thus, to compute the hypergraph cut in a sense of~\cite{ZhouHyper,saito2018hypergraph,ghoshdastidar2015provable}, this can be computed in $O(n^3)$.
If we go further for the case where $m$ is odd, we also lose the kernel sense; we cannot obtain the Eq.~\eqref{eq:matrixgeneralizedkernel}-like a good representation of the contracted matrix.
Thus, we need a naive computational time, $O(n^{m})$.

The discussion above shows how we lose the kernel property if we expand from a half-symmetric kernel to a symmetric kernel. 
In this sense, for the standard case, we use the half-symmetricity even though if $m=2$ when a tensor is half-symmetricity, then symmetricity.

In graph research, the Gaussian modeling is theoretically connected the standard graph cut Eq.~\eqref{eq:kwaynormalized}, as discussed in the extended Related Work in the Appendix. 
However, we often see the Gaussian modeling is applied to the different cut objective function than the standard spectral clustering ones.
In the same sense, the following is a possible scenario; we only use the modeling and use the different cut objective function than Eq.~\eqref{eq:defhypergraphcut}, which is used in~\cite{ZhouHyper,ghoshdastidar2015provable,saito2018hypergraph}.
The problem for this use is that our representation is only for even $m$; thus, even for the smallest $m$-uniform hypergraph $m=4$, it costs expensive $O(n^4)$ to create a hypergraph.
Seeing the discussion in this section, introducing symmetricity, or even reducing $m=3$ needs some sacrifice of the theoretical support.
The main theoretical supports that we lose is from the heat kernel discussion Thm.~\ref{thm:lapalcianconverge}.
Hence, unless we need a symmetric modeling, we recommend using sampling form a hypergraph by using the original biclique kernel Eq.~\eqref{eq:generalizedkernel}.
The sampling technique is also used in~\cite{ghoshdastidar2015provable,li2018submodular}.

\subsection{Inhomogeneous Cut}

An inhomogeneous hypergraph cut is a cut objective which assigns different costs to the different cuts of edges~\cite{li2017inhomogoenous,veldt2020hypergraph}. 
More formally, the 2-way cut Eq.~\eqref{eq:defhypergraphcut} can be rewritten for inhomogeneous cut as
\begin{align}
\label{eq:definhcut}
    \mathrm{Cut}(V_{1}, V_{1}\backslash V) = \sum_{e \in E} w_{e}(e \cap V_{1}),
\end{align}
where $w_{e}$ is called as \textit{splitting function} for edge $e$ and a split of $e$ incurred by $V$. 
This extends to $k$-way cut in the same way as Eq.~\eqref{eq:kwaypartitioningmuniformhypergraph}. 
Assume that $w_{e}(e\cap V)$ is submodular and only depends on cardinally, i.e., depends only on $|V|$. 
If we change in the objective Eq.~\eqref{eq:defgeneralizedkernelkmeans} as
\begin{align}
    \kappa^{(m)}(i,i_{\cdot},l,l_{\cdot}) \to \kappa^{(m)}_{\mathrm{inh}}(i,i_{\cdot},l,l_{\cdot};V_{i}),
\end{align}
where $\kappa^{(m)}_{\mathrm{inh}}$ serves as a cost of splitting function for a multi-way modeling for $w_{e}(e \cap V)$, the weighted kernel $k$-means objective style discussion can be connected to the inhomogeneous cut Eq.~\eqref{eq:definhcut}. 
However, since $\kappa^{(m)}_{\mathrm{inh}}$ is \textit{not} a biclique kernel in general, we cannot apply the other theoretical results in this paper to $\kappa^{(m)}_{\mathrm{inh}}$.

We rewrite the Eq.~\eqref{eq:definhcut} as
\begin{align}
\notag
    &\mathrm{Cut}(V_{1}, V_{1}\backslash V) = \\
\notag
    &\sum_{v,u \in V} \sum_{v_{\cdot},u_{\cdot}, V} w_{e}(e=(v,v_{\cdot},u,u_{\cdot}) \cap V_{1})
\end{align}
Using the results from~\cite{veldt2020hypergraph}, if $w_{e}(e\cap V)$ is submodular and only depends on cardinally, i.e., depends only on the number of vertices in one set, 
\begin{align}
\notag
   &\mathrm{Cut}(V_{1}, V_{1}\backslash V)  = \\
   \label{eq:defcardinarity}
   &\sum_{v\in V_{1}, u \in V \backslash V_{1}} \sum_{v_{\cdot},u_{\cdot}, V}  w_{e}(e=\{v,v_{\cdot},u,u_{\cdot}\} \cap V_{1}).
\end{align}
\citet{veldt2020hypergraph} conjectured that this holds for \textit{any} submodular splitting functions, since there is a good ``sense'' to believe that this is true (Cor.~4.3 in~\cite{veldt2020hypergraph}). 
We refer to~\citet{veldt2020hypergraph} for the detailed discussion. 
Now we put as
\begin{align}
\label{eq:definhgeneralizedkernel}
 \kappa^{(m)}_{\mathrm{inh}}(i,i_{\cdot},l,l_{\cdot};V_{i}) := w(\{i,i_{\cdot},l,l_{\cdot}\} \cap V_{i}),
\end{align}
and we define the contraction of inhomogeneous weight as
\begin{align}
 K^{(m)}_{\mathrm{inh}}(i,j;V_{i}) := \sum_{i_{\cdot},l_{\cdot}}w(\{i,i_{\cdot},l,l_{\cdot}\} \cap V_{i}).
\end{align}
Now we obtain
\begin{align}
\notag
    &J'(\{\pi_j\}_{j=1}^{k}) =
    \sum_{i \in \pi_{j},j}
    \sum_{
    \{i_{\cdot}\}
    \subset \pi_{j}
    }
    w_{i}\kappa^{(m)}( i,i_{\cdot},i,i_{\cdot}; \pi_{j}) \\ 
    \notag
    &
    -\sum_{i,l \in \pi_j,j}
    \sum_{
    \{i_{\cdot}\}, \{l_{\cdot}\} \subset \pi_{j}
    } w_{i}w_{l}\kappa^{(m)}(i,i_{\cdot},l,l_{\cdot};\pi_{j})/s_{j}\\
    \label{eq:matrixforminh}
    &= W^{1/2} K^{(m)}_{\mathrm{inh}} W^{1/2} - Y^{\top}W^{1/2} K^{(m)}_{\mathrm{inh}} W^{1/2}Y.
\end{align}
From Eq.\eqref{eq:defcardinarity}, this is connected to the definition of inhomogeneous cut Eq.~\eqref{eq:definhcut}.
Observing Eq.~\eqref{eq:matrixforminh}, if $K^{(m)}_{\mathrm{inh}}(i,j;V_{i})$ forms a kernel, this can be written Eq.~\eqref{eq:kmeansobjectiveforbicliquekernel}, and therefore we can apply $k$-means style algorithm.
If $\kappa^{(m)}_{\mathrm{inh}}(i,i_{\cdot},l,l_{\cdot};V_{i})$ is biclique kernel, we can apply Thm.~\ref{thm:kernelsemidefinite} and Thm.~\ref{thm:lapalcianconverge}.
On top of the conjecture from~\cite{veldt2020hypergraph}, it is still open problem which class of splitting function has kernel properties as discussed above.

We conducted some preliminary experiments for the inhomogeneous cut; see the next section.

\begin{table*}[!t]
\small
\centering
\begin{tabular}{c|cccc}
Kernel and Method                                                          & iris                    & spine                   & ovarian                 & Hopkins155                      \\\hline\hline
Gaussian ($m$$=$$2$, the standard graph)                                 & 6.99E-05$\pm$3.54E-06   & 4.05E-04$\pm$1.11E-09   & 2.18E-04$\pm$7.29E-10   & 4.82E-02 \\
Gaussian Ours ($m$$\geq$$4$)                              & 6.05E-05$\pm$2.90E-06   & 4.28E-04$\pm$1.13E-09   & 2.41E-04$\pm$6.56E-10   & 5.22E-02                     \\
Gaussian~\cite{ghoshdastidar2015provable}            & 1.57E-03$\pm$4.44E-06   & 1.54E-02$\pm$3.15E-06   & 4.31E-03$\pm$1.45E-06   & 3.69E+00                     \\
Gaussian (Affine Supspace)                                                 & 9.16E-04$\pm$1.06E-05   & 7.14E-03$\pm$5.51E-06   & 2.70E-02$\pm$6.11E-06   & 4.12E+01                     \\
Gaussian~($d^{H-2}$~\cite{li2017inhomogoenous}) & 6.12E-02$\pm$2.54E-06 & 2.15E-01$\pm$1.08E-06 & 1.01E-01$\pm$3.15E-06 & 1.22E+01 \\\hline
Polynomial ($m$$=$$2$, the standard graph)                               & 3.94E-05$\pm$6.98E-06   & 1.11E-04$\pm$2.18E-09   & 7.29E-05$\pm$4.82E-07   & 2.69E-02 \\
Polynomial Ours ($m$$\geq$$4$)                            & 3.11E-05$\pm$3.73E-06   & 6.73E-05$\pm$2.17E-09   & 1.07E-04$\pm$4.96E-05   & 2.20E-02 \\
Polynomial~\cite{ghoshdastidar2015provable}                                                      & 1.57E-03$\pm$4.44E-06   & 1.54E-02$\pm$3.15E-06   & 4.31E-03$\pm$1.45E-06   & 3.69E+00                     \\
Polynomial~\cite{yu2018modeling}    & 9.16E-04$\pm$1.06E-05   & 7.14E-03$\pm$5.51E-06   & 2.70E-02$\pm$6.11E-06   & 4.12E+01                    
\end{tabular}
\caption{Runtime Summary (unit:secs). Here we use E notation, e.g., E-06
$=10^{-6}$. For Hopkins155, we sum up all the computational time, and report the time which produced the best error result summarized in Table~\ref{tab:results}}
\label{tab:exp_times}
\end{table*}

\begin{figure*}[!t]
\begin{center}
\subfigure[Iris]{%
\includegraphics[width=.24\hsize,clip]{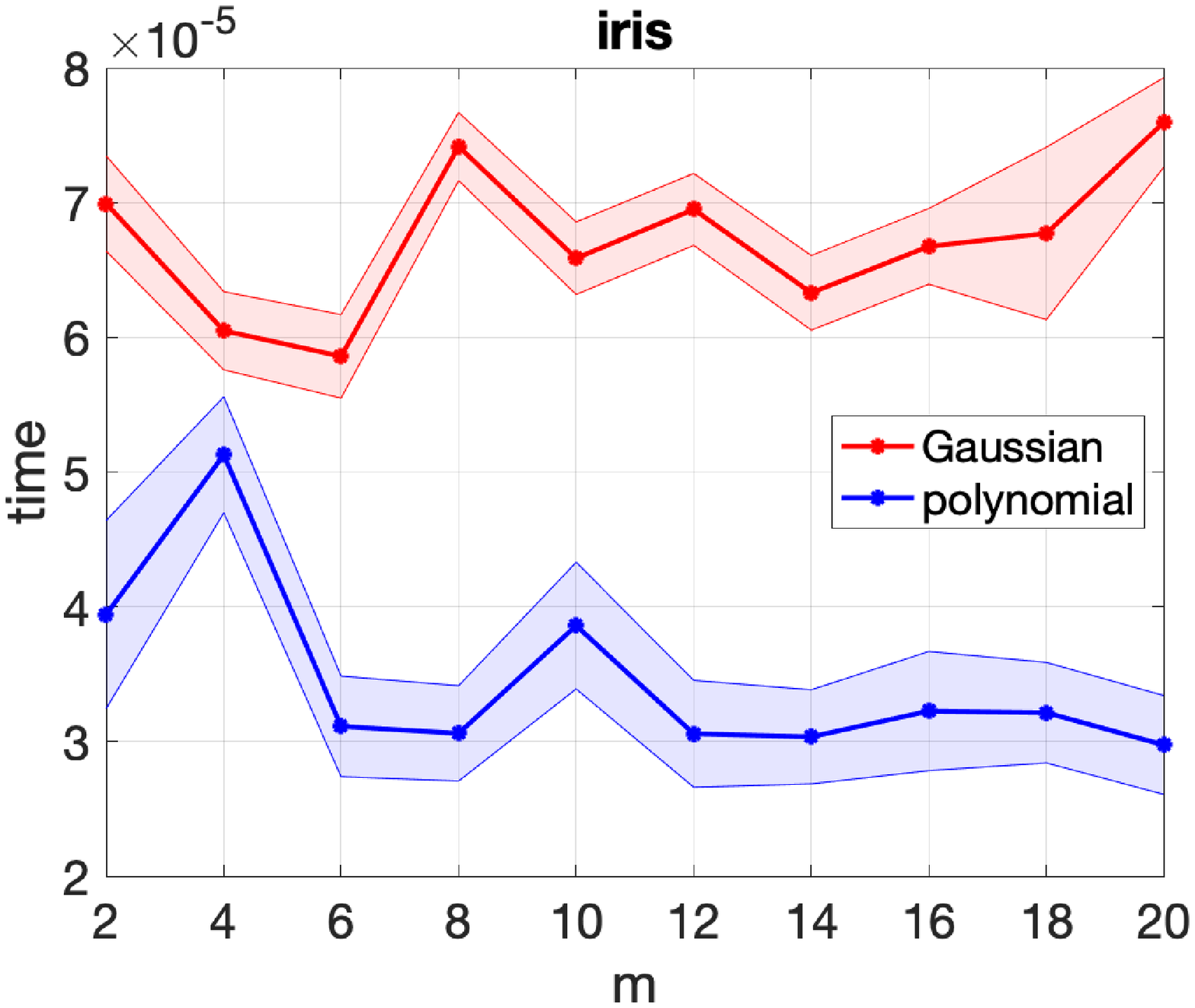}\label{fig:iris_times}}
\subfigure[Spine]{%
\includegraphics[width=.24\hsize,clip]{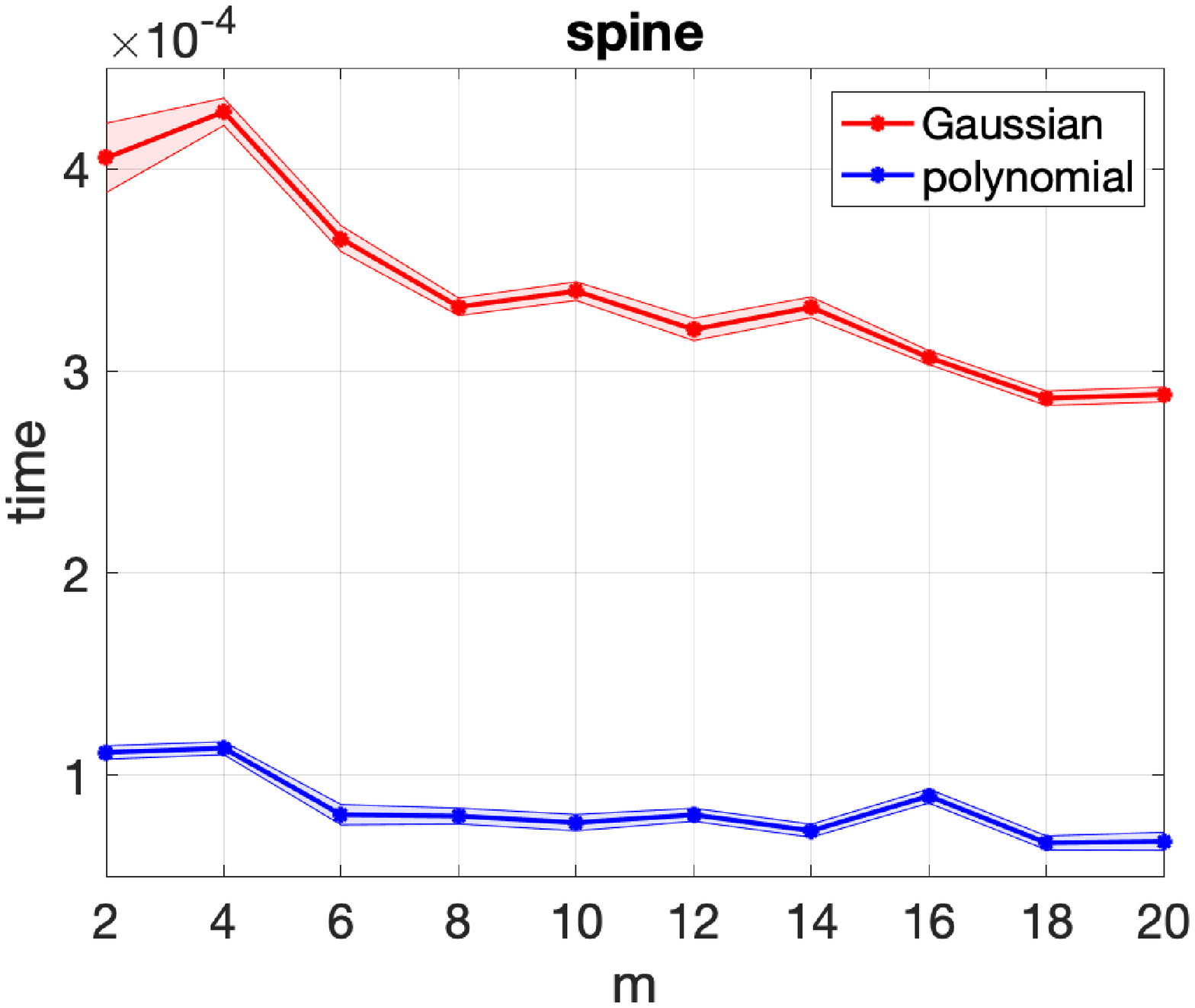}\label{fig:spine_times}}
~\subfigure[Ovarian]{%
\includegraphics[width=.24\hsize,clip]{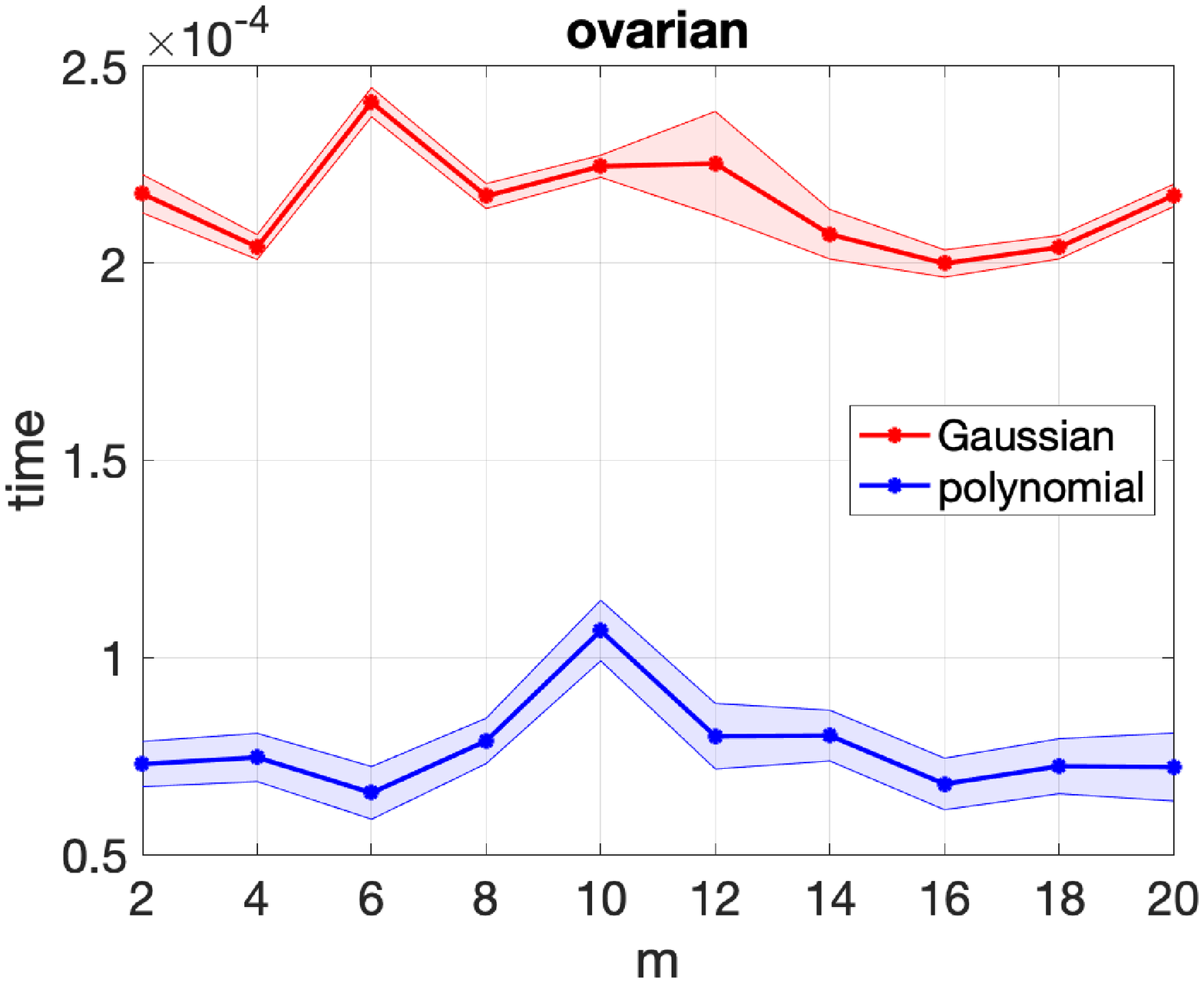}\label{fig:ovarian_times}}
~\subfigure[Hopkins155]{%
\includegraphics[width=.23\hsize,clip]{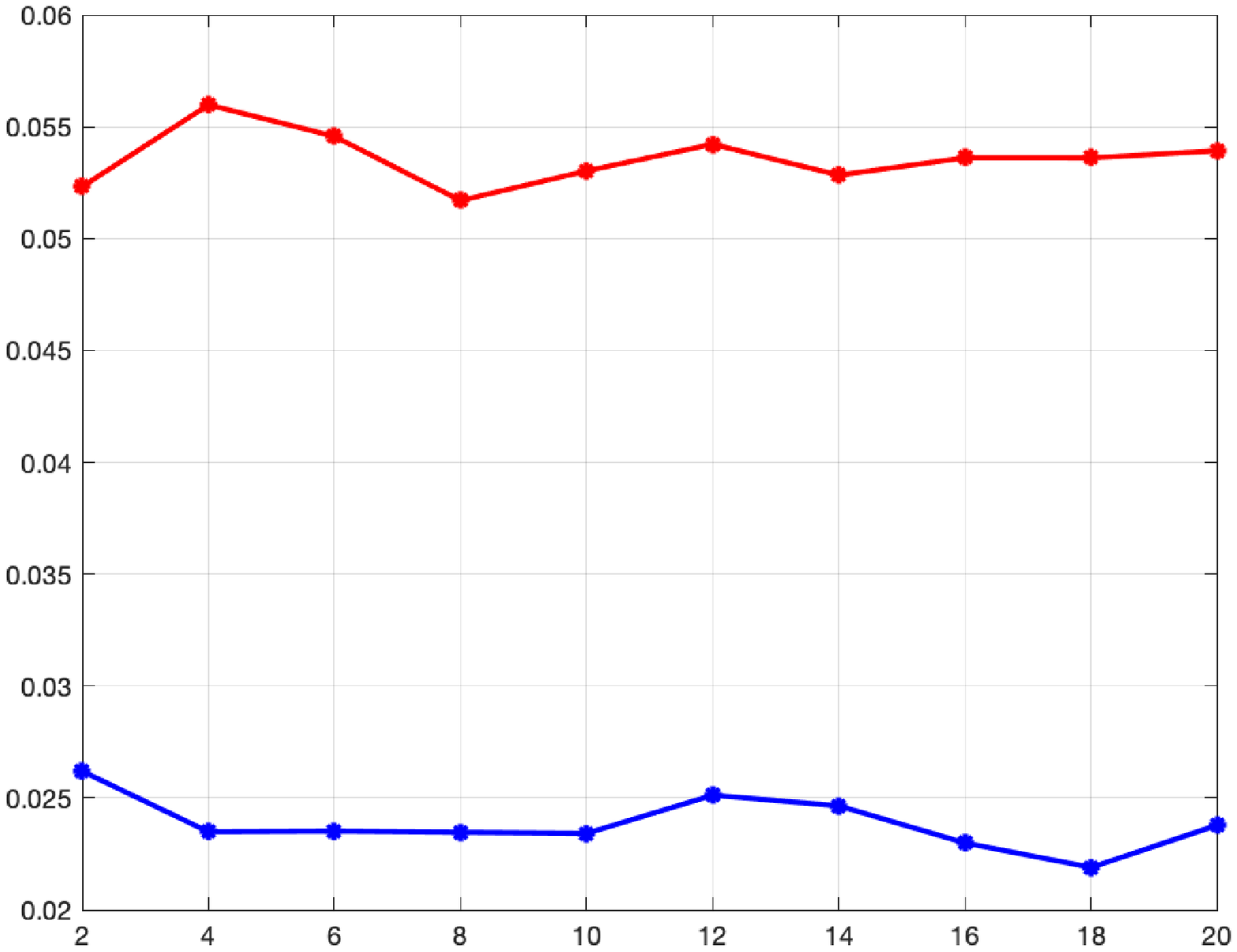}\label{fig:motion_times}}
\caption{Runtime for our methods. Red shows the result for Gaussian and blue shows for polynomial. The shade shows the standard deviation of the fourth step of Alg.~\ref{algo:spectralclustering}. Since Hopkins155 is the sum of runtime of different 155 datasets, this only shows the average.}
\label{fig:exp_time}
\end{center}
\end{figure*}

\begin{table*}[!t]
\centering
\begin{tabular}{c|cccc}
        & iris & spine& ovarian& Hopkins155 \\
        \hline\hline
Ours with inhomogeneous (Eq.~\eqref{eq:onevariableinh}) & 0.4822$\pm$0.0113 & 0.3284$\pm$0.0067& 0.3952$\pm$0.0000& 0.4000\\
inhomogeneous $d^{\mathrm{inh}}$~\cite{li2017inhomogoenous} & 0.0853$\pm$0.0645 & 0.3026$\pm$0.0539&0.1111$\pm$0.0000&\textbf{0.0732}\\
Ours $m=4$ & 0.0723$\pm$0.0319 & \textbf{0.2806$\pm$0.000} & \textbf{0.0841$\pm$0.0000} & 0.1112\\
Ours best & \textbf{0.0697$\pm$0.0033} & \textbf{0.2806$\pm$0.000} &  \textbf{0.0841$\pm$0.0000} & 0.0927                        
\end{tabular}
\caption{Additional Experimental Results. The standard deviation is from randomness involved in the fourth step of Alg.~\ref{algo:spectralclustering}. Since Hopkins155 is the average performance of 155 datasets, this only shows the average.}
\label{tab:exp_resutlsadd}
\end{table*}

\section{Additional Experimental Results.}
\label{sec:additionalexperimental}

\begin{table}[!t]
\centering
\begin{tabular}{c|cccc}
          & iris & spine & ovarian  & hopkins155\\\hline\hline
size      & 150 & 310 & 216 & 45850 \\
dimension & 4 & 6  & 100 & 89.32
\end{tabular}
\caption{Dataset Summary. Since Hopkins 155 contains 155 different videos, we report the sum of the data points and average dimensions of videos.}
\label{tab:datasetsummary}
\end{table}

We here give more description on comparison methods. 
Also, we give more detailed experimental results, runtime. 
Finally, we demonstrate the preliminary experiments on inhomogeneous cut.

Firstly, we give a details of dataset in Table~\ref{tab:datasetsummary}.
Next, we give a details of the compariosn methods. We used an ad-hoc modeling used in~\cite{ghoshdastidar2015provable} for both kernels, which is $\mathcal{A}_{ijk}$$:=$$\max$$(\kappa(\mathbf{x}_{i}$$,\mathbf{x}_{j})$$,\kappa(\mathbf{x}_{j},$$\mathbf{x}_{k})$$,\kappa(\mathbf{x}_{k},\mathbf{x}_{i}))$.
Thirdly, we employ a Gaussian-type modeling used in computer vision~\cite{govindu2005tensor}, which is $\mathcal{A}_{ijk}$$:=$$\exp (- \gamma \lambda_1)$, where $\lambda_1$ is the smallest eigenvalue of $\mathbf{X}_{ijk}^{\top}\mathbf{X}_{ijk}$, and $\mathbf{X}_{ijk}$$:=$$(\mathbf{x}_i,\mathbf{x}_j,\mathbf{x}_k)$. 
The $d^{H-2}$ is an Euclidean distance between $v$ and the affine subspace generated by $e/\{v\}$, for all $v \in e$, and sum this up for all $v \in e$.
Lastly, we used a generalized dot product form~\cite{yu2018modeling}, that is $\mathcal{A}_{ijk}$$:=$$\sum_{l} x_{il}x_{jl}x_{kl}$.
We would like to mention that our experiments was run with Matlab on Mac Mini with Intel i7 Processor and 32GiB RAM.

Although all the comparisons and ours equally cost $O(n^3)$, we provide the runtime of our experiment in Table~\ref{tab:exp_times} and Fig.~\ref{fig:exp_time}.
Ours are faster than the hypergraph comparisons. This comes from the difference in construction of hypergraphs. All methods have roughly two parts; i) construction of hypergraph and ii) spectral clustering, and ii) costs $O((n^3))$
 in all the methods. 
 However, for the construction of hypergraphs, while ours costs $O(n^2)$, the other comparisons cost $O(n^3)$ to construct. This difference induces the time difference.
The actual running time of our method does not change the time very much when we increase the order of the hypergraph $m$. 
This supports our claim -- no matter which $m$ we take, the overall computational time does not depend on $m$. 
On the other hand, for the comparison methods, if we increase $m$ we expect the actual running time to increase since the comparisons cost $O(n^m)$ to compute.

Finally, we demonstrate a preliminary experiment on inhomogeneous cuts. 
This experiment aims to try an inhomogeneous cut objective for our formulation and compare it with existing methods. 
For inhomogeneous cut, we apply the approximation algorithm proposed by~\cite{li2017inhomogoenous}.
The algorithm needs the ``effect'' of one vertex to the edge.
Using the symmetric modeling Eq.~\eqref{eq:defsymmetric}, a naive but straightforward formulation of such a splitting function for the edge $e=\{\mathbf{x}_{i_1},\ldots,\mathbf{x}_{i_m}\}$ and one vertex $v_{i_{\mu}}$ can be formulated as
\begin{align}
\label{eq:onevariableinh}
    w_{e} (\{v=\mathbf{x}_{i_\mu}\}) := \sum_{j \in \{1,\ldots,m\}\backslash \{\mu\}} \kappa(\mathbf{x}_{i_{\mu}}, \mathbf{x}_{i_{j}}).
\end{align}
Then, we apply the algorithm by~\cite{li2017inhomogoenous} to Eq.~\eqref{eq:onevariableinh}.

The experimental settings are as follows.
As a comparison, we use $d^{\mathrm{inh}}$ used in~\cite{li2017inhomogoenous}, which is the Euclidean distance between $v$ and the affine subspace generated by $e/\{v\}$, for all $v \in e$. 
Note that $d^{H-2} = \sum_{v \in e}d^{\mathrm{inh}}(\{v\})$.
Since the existing modeling $d^{\mathrm{inh}}$ is Gaussian-type, we only compare with Gaussian modeling. 
Moreover, since Eq.~\eqref{eq:onevariableinh} only can be defined for even $m$, we employ $m=4$ for Eq.~\eqref{eq:onevariableinh} as well as $d^{\mathrm{inh}}$.
We random sample the edges in the same procedure as~\cite{li2017inhomogoenous} since it is too time-consuming if we take $O(n^4)$ to compute hypergraph edge.
We also compare the best performance of our biclique kernel modeling and ours with $m=4$. 
The rest of the procedure is the same as the main experiment. 
For Hopkins155, we report the performance of $m=4$, and we also report the average of the best performance of each dataset.

The experimental result is summarized in Table~\ref{tab:exp_resutlsadd}.
For iris, spine, ovarian, ours outperformed the others, especially $d^{\mathrm{inh}}$, which is believed to be a more sophisticated splitting function than the cut function we use.
This might come from the benefit of using the formulation of multi-way similarity, which is theoretically connected to the hypergraph cut algorithm.
While our formulation and cut algorithm are theoretically connected, the formulation of $d^{\mathrm{inh}}$ is not theoretically supported by the algorithm from~\cite{li2017inhomogoenous}. 
However, a naive but straightforward inhomogeneous formulation of ours, Eq.~\eqref{eq:onevariableinh} poorly worked. 
This result suggests that modeling for an inhomogeneous splitting function is not a silver bullet -- we need to consider a proper splitting function for each case. Moreover, suitable modeling for this approximated inhomogeneous cut algorithm is different from our biclique kernel formulation.
This experiment suggests that we need to try more splitting functions to obtain a comprehensive view of the modeling that matches inhomogeneous cut as future work. 
It would also be nice if we could explore some splitting function and modeling that is theoretically connected to the algorithm of~\cite{li2017inhomogoenous}.

\end{document}